%% file: arxiv.tex
\documentclass[runningheads]{llncs}

\usepackage[mobile]{eccv}

\usepackage{eccvabbrv}
\usepackage{graphicx}
\usepackage[accsupp]{axessibility}
\usepackage{bm}
\usepackage{amsfonts} 
\usepackage{textcomp}
\usepackage{stmaryrd}
\usepackage[ruled,vlined]{algorithm2e}
\input{macros}

\DeclareMathOperator*{\argmin}{argmin}

\usepackage[accsupp]{axessibility}  

\usepackage{hyperref}

\usepackage{orcidlink}

\begin{document}

\title{Gravity-aligned Rotation Averaging with\\ Circular Regression} 


\author{Linfei Pan$^1$
\and Marc Pollefeys$^{1,2}$
\and Dániel Baráth$^1$\\
$^1$ETH Zurich
\quad\quad $^2$Microsoft\\
\texttt{\{linfei.pan, marc.pollefeys, danielbela.barath\}@inf.ethz.ch}
}

\authorrunning{L.~Pan et al.}

\institute{
}

\maketitle

\begin{abstract}
    Reconstructing a 3D scene from unordered images is pivotal in computer vision and robotics, with applications spanning crowd-sourced mapping and beyond. While global Structure-from-Motion (SfM) techniques are scalable and fast, they often compromise on accuracy. To address this, we introduce a principled approach that integrates gravity direction into the rotation averaging phase of global pipelines, enhancing camera orientation accuracy and reducing the degrees of freedom. This additional information is commonly available in recent consumer devices, such as smartphones, mixed-reality devices and drones, making the proposed method readily accessible. Rooted in circular regression, our algorithm has similar convergence guarantees as linear regression. It also supports scenarios where only a subset of cameras have known gravity. Additionally, we propose a mechanism to refine error-prone gravity. We achieve state-of-the-art accuracy on four large-scale datasets. Particularly, the proposed method improves upon the SfM baseline by 13 AUC@$1^\circ$ points, on average, while running eight times faster. It also outperforms the standard planar pose graph optimization technique by 23 AUC@$1^\circ$ points. The code is at \url{https://github.com/colmap/glomap}.
\end{abstract}

\input{tex/text/introduction}

\input{tex/text/related_work}

\input{tex/text/method}

\input{tex/text/experiments}
\input{tex/text/conclusion}

\section*{Acknowledgment}
The authors thank Mihai Dusmanu and Paul‑Edouard Sarlin for their help on the LaMAR dataset.
We also want to thank Yunke Ao and Shaohui Liu for helping to review the paper.
This work was partially funded by the Hasler Stiftung Research Grant via the ETH Zurich Foundation and the ETH Zurich Career Seed Award.
Linfei Pan was supported by gift funding from Microsoft.

%
%
\bibliographystyle{splncs04}
\bibliography{main}

\newpage
\begin{center}
\textbf{\large Supplemental Materials}
\end{center}
\setcounter{equation}{0}
\setcounter{figure}{0}
\setcounter{table}{0}
\setcounter{section}{0}
\makeatletter
\renewcommand{\theequation}{S\arabic{equation}}
\renewcommand{\thefigure}{S\arabic{figure}}
\renewcommand{\thetable}{S\arabic{table}}
\renewcommand{\thesection}{S\arabic{section}}

\input{tex/text/appendix}

\end{document}

%% file: macros.tex
\usepackage{booktabs}
\usepackage{balance}
\usepackage{multirow}
\usepackage{overpic}
\usepackage{cite}
\usepackage[font=small,labelfont=bf]{caption}
\usepackage{color}
\usepackage{pifont}
\usepackage{tikz}
\usetikzlibrary{patterns}

\usepackage{colortbl}
\usepackage{pgfplots}
\pgfplotsset{compat=1.17}
\usepackage{tabularx}
\usepackage{arydshln}
\usepackage{amssymb}
\usepackage{pifont}

 \usepackage{amsmath}
\usepackage{wasysym}%
\usepackage{enumitem}
\usepackage{wrapfig}
\usepackage{xspace}
\usepackage{tabu}
\usepackage[super]{nth}
\usepackage{subcaption}
\usepackage{scalefnt}
\usepackage{dsfont}
\usepackage{graphicx,calc}

\definecolor{darkgreen}{RGB}{0,153,51}
\definecolor{linkgreen}{RGB}{52,130,48}
\definecolor{LightCyan}{rgb}{0.87,0.92,0.96}
\definecolor{m_green}{RGB}{233, 254, 187}
\definecolor{m_orange}{RGB}{255, 212, 121}
\definecolor{m_red}{RGB}{255, 190, 188}
\definecolor{m_violet}{RGB}{215, 131, 255}
\definecolor{m_blue}{RGB}{186, 234, 255}
\definecolor{m_brown}{RGB}{255,212,120}
\definecolor{m_lightgreen}{RGB}{212,251,122}
\definecolor{notetext}{rgb}{0.7,0,0}

\definecolor{model_pink}{RGB}{235, 106, 164}
\definecolor{model_orange}{RGB}{250, 194, 122}
\definecolor{model_green}{RGB}{164, 210, 162}
\definecolor{model_gray}{RGB}{120, 120, 120}
\definecolor{model_yellow}{RGB}{251, 231, 171}
\definecolor{model_purple}{RGB}{190, 146, 211}

\def\eg{\emph{e.g.}\@\xspace} 
\def\ie{\emph{i.e.}\@\xspace}

\renewcommand{\vec}[1]{\ensuremath{\mathbf{#1}}}

%% file: tex/text/introduction.tex
\section{Introduction}

Reconstructing a 3D model from a large set of unordered images is a critical task in computer vision and robotics, with numerous applications in crowd-sourced mapping and other domains. 
The most popular approach for 3D reconstruction is the \textit{Structure-from-Motion} algorithm, which estimates both the camera parameters and 3D point cloud simultaneously~\cite{ullman1979interpretation}. 
Work in this field generally falls into two categories: \textit{Incremental} \cite{heinly2015reconstructing, schoenberger2016sfm, snavely2006photo, snavely2008modeling, wu2013towards} and \textit{Global methods} \cite{cui2015global, zhu2018largesfm, theia-manual}. 
Incremental methods proceed by carefully selecting and adding images one-by-one to the 3D reconstruction while maintaining accuracy through numerical optimization.
Even though such methods lead to particularly accurate results, they can be computationally prohibitive due to the need for frequent bundle adjustment \cite{triggs2000bundle}.
Also, they do not scale well to large image collections. 
On the other hand, global methods directly optimize a noisy pose graph and consider all cameras simultaneously, resulting in a faster and more scalable approach.
Although global methods are generally considered less accurate than their incremental counterparts \cite{cui2017hsfm}, they provide a promising research direction for lightweight 3D reconstructions that run in a matter of minutes. 
This paper focuses on global algorithms and proposes an algorithm to improve camera rotation estimation by incorporating the gravity direction as additional sensory information into the procedure.
This information is available \textit{by default} in recent consumer devices such as smartphones or mixed reality devices, making it a potentially useful and widely accessible data source.

\begin{figure}[t]
    \centering
    \includegraphics[width=1.0\columnwidth]{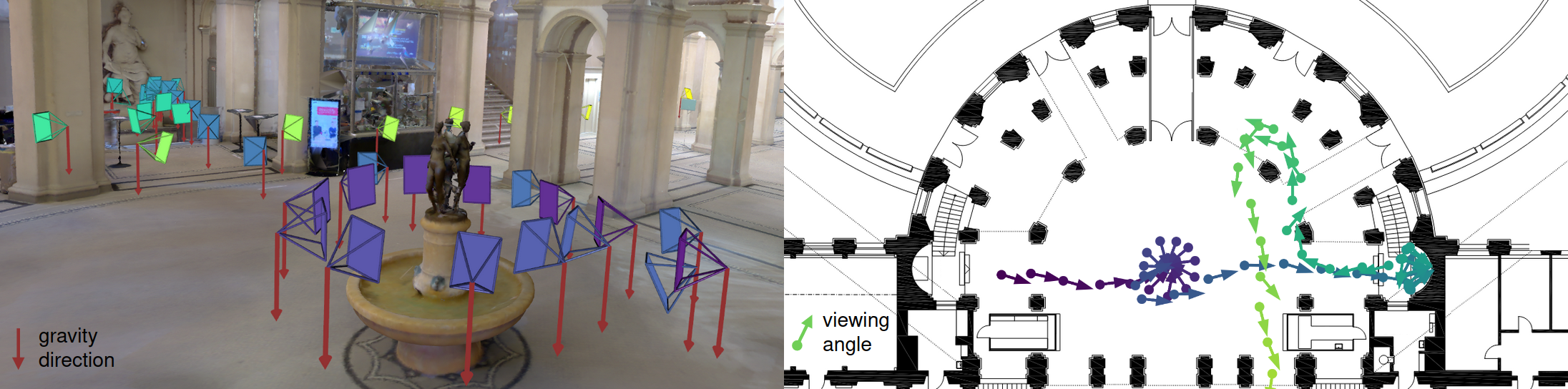}
    \caption{Gravity-aligned rotation averaging. 
    Given a set of cameras and their gravity directions (left), orientation estimation becomes a 1-degree-of-freedom optimization of the viewing angle (right). 
    Circular regression tackles the periodicity of the problem.
    }
    \label{fig:visualization}
\end{figure}

Generally, global methods start with selecting image pairs with a common field-of-view employing image retrieval, such as the visual bag-of-words algorithm \cite{filliat2007visual}. 
Then, relative pose estimation is performed between the pairs using a robust estimator, \eg, RANSAC \cite{fischler1981random}.
The resulting relative poses are used to construct an initial pose graph, and camera rotations are estimated simultaneously by \textit{rotation averaging} \cite{govindu2001combining, hartley2013rotation}. 
Given the estimated camera rotations, translations and a 3D point cloud are estimated \cite{wilson2014robust}.
Finally, bundle adjustment is performed to refine the 3D reconstruction and camera parameters. 

Rotation averaging, also called rotation synchronization, has a rich research literature~\cite{hartley2013rotation, martinec2007robust} and is an integral part of pose graph optimization (PGO) algorithms~\cite{carlone2012linear, wu2023decoupled}.
Typically, it is formulated as a non-linear optimization over the consistency of estimated relative poses and unknown global ones, with the rotation matrices represented by their three degrees of freedom (DoFs).
However, modern devices like smartphones, drones and robots, are equipped with inertial measurement units (IMUs), enabling us to pre-estimate the gravity direction accurately~\cite{madgwick2011estimation}. 
Such data is widely used in, \eg,
autonomous driving \cite{liu2019edge}, unmanned aerial vehicles control \cite{feng2021comprehensive}, and navigation \cite{chang2020gnss, petovello2003real}.
Having a common direction can resolve the yaw and roll of the camera rotation, reducing 2 DoFs in the orientation estimation.
Similar simplification applies to rotation averaging, and consequently, the problem becomes a 1-DoF optimization and aligns with the planar PGO problems \cite{carlone2012linear, carlone2018convex}, as visualized in Fig.~\ref{fig:visualization}.

Planar pose graph optimization assumes that cameras move in a plane, necessitating only the estimation of the viewing direction. 
LAGO~\cite{carlone2014angular}, a representative method in rotation estimation for planar PGO, recognizes that the single degree of freedom permits almost linear estimation, though the periodicity of the variable introduces non-linearity.
To address this, LAGO pre-calculates periods via a heuristic approach and keeps it constant for the subsequent steps. 
While many planar PGO methods adopt this approach \cite{carlone2012linear, wu2023decoupled}, the fixed periodicity makes them vulnerable to outliers and error-prone initialization.

The main contribution of this work is a new principled solution to planar rotation averaging.
Unlike prior work~\cite{carlone2012linear, wu2023decoupled} that fix periods and solve linear systems, we tackle the problem via circular regression.
Combined with robust losses, it leads to significantly more accurate results on four large-scale and real-world datasets (EuRoC~\cite{burri2016euroc}, KITTI~\cite{geiger2013vision}, LaMAR~\cite{sarlin2022lamar}, and 1DSfM~\cite{wilson2014robust}). 
Our improvements also extend to the case when only a subset of cameras have known gravity.
We further introduce a refinement method for gravity.

\begin{figure*}[t]
    \centering
        \subfloat[EuRoC \cite{burri2016euroc}]{%
         \includegraphics[height=0.33\textwidth]{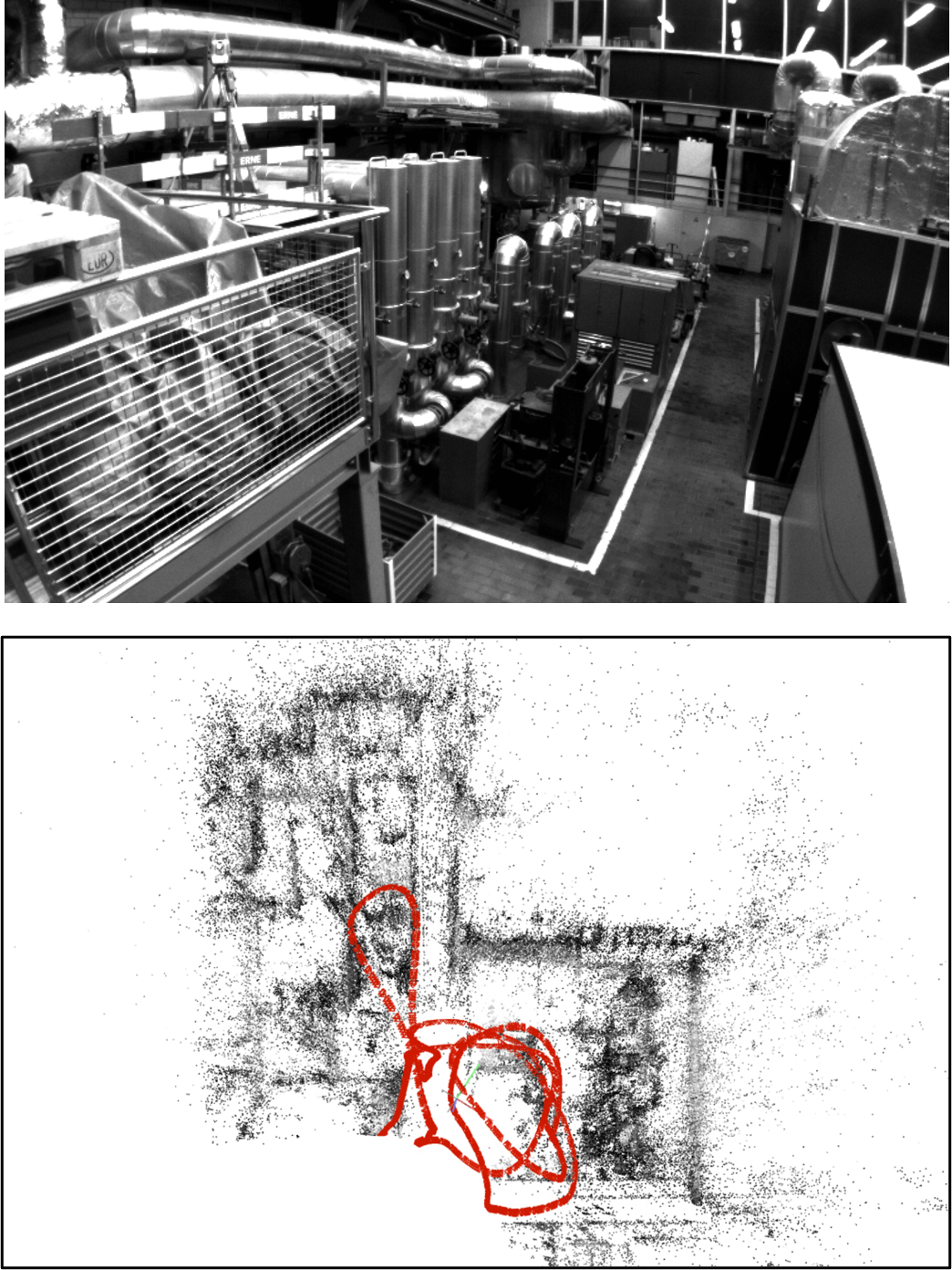}
         \label{fig:examplary_euroc}
         }
        \subfloat[KITTI \cite{geiger2013vision}]{%
         \includegraphics[height=0.33\textwidth]{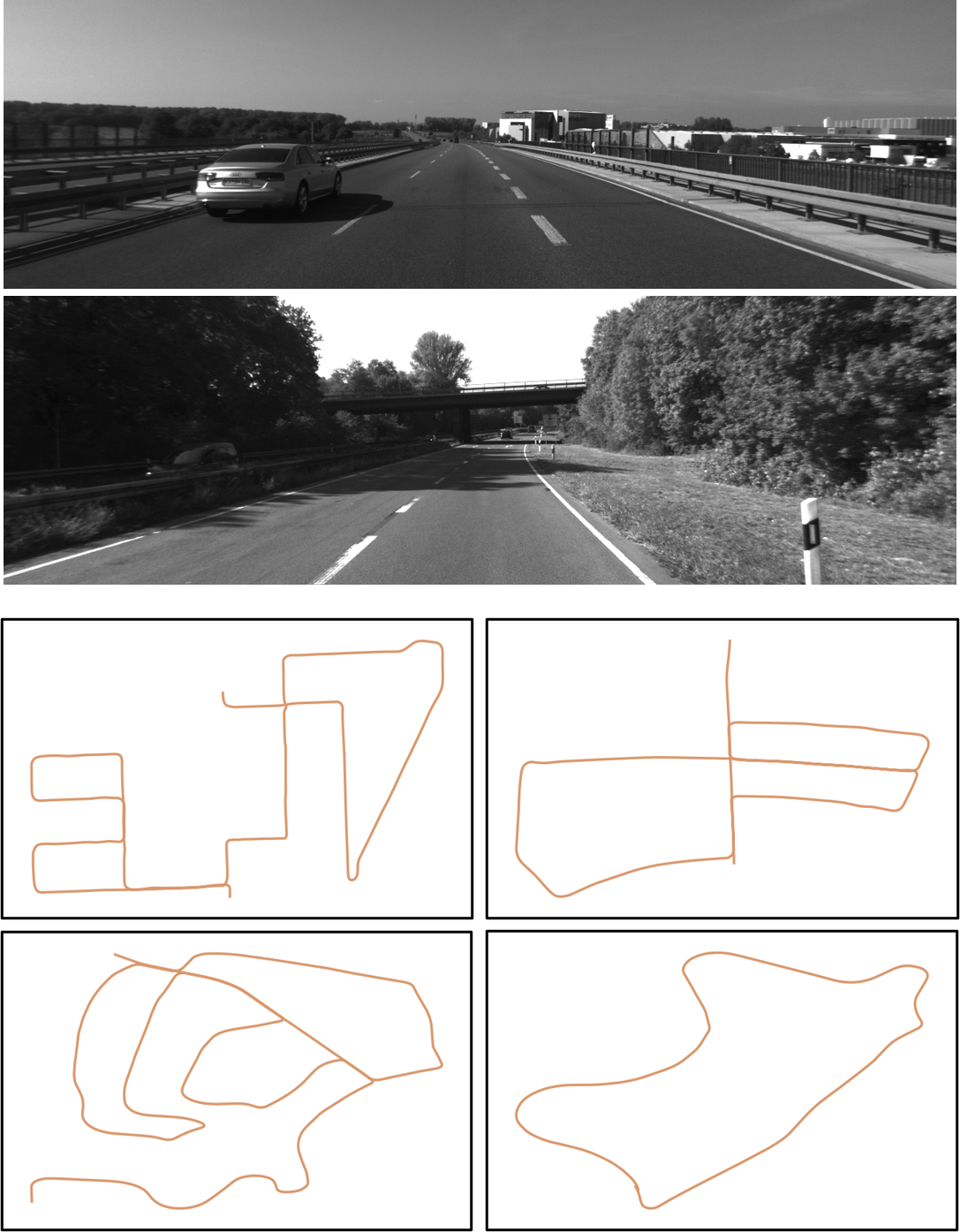}
         \label{fig:examplary_kitti}
         }
        \subfloat[LaMAR \cite{sarlin2022lamar}]{%
         \includegraphics[height=0.33\textwidth]{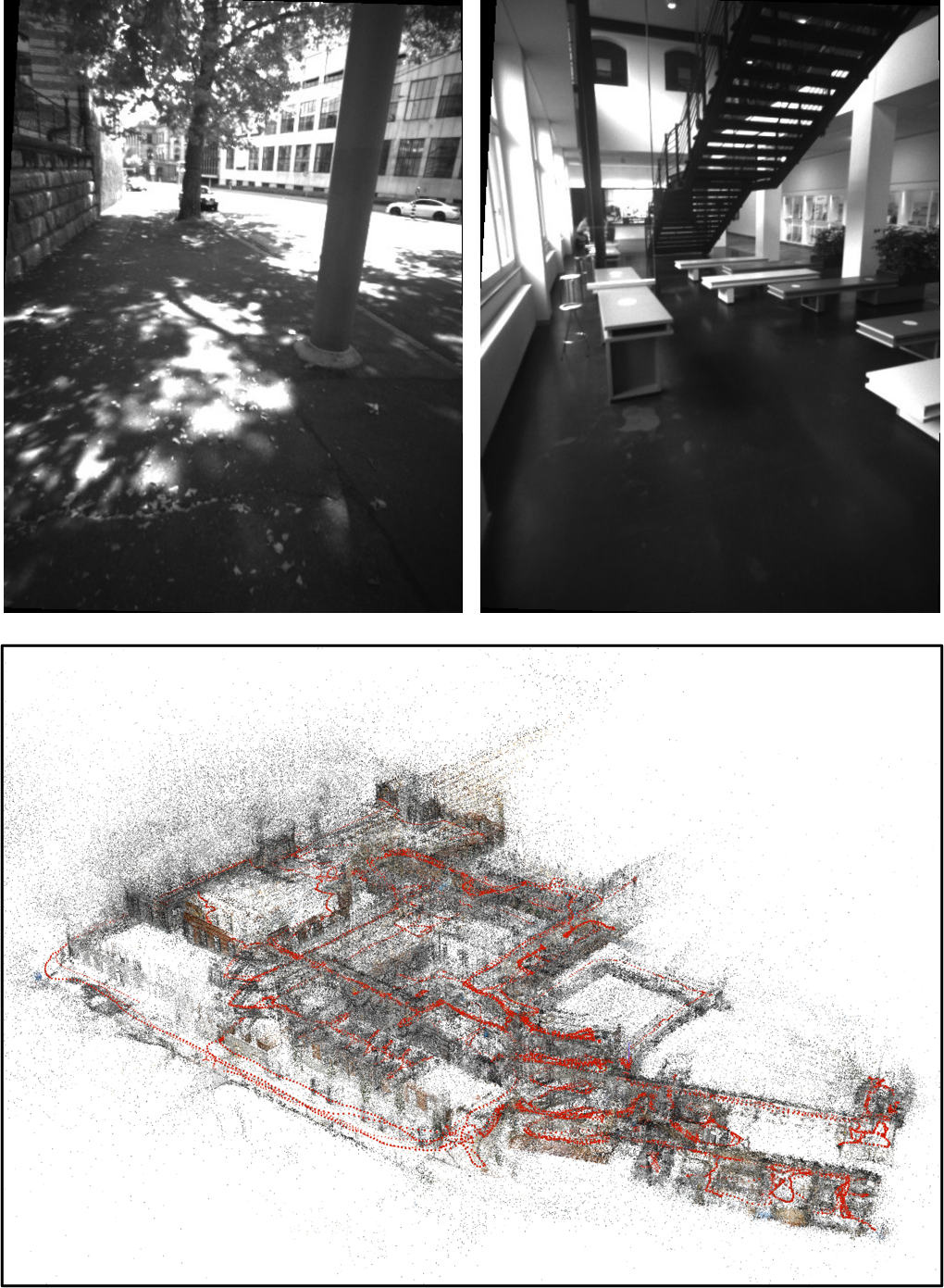}
         \label{fig:examplary_lamar}
         }
        \subfloat[1DSfM \cite{wilson2014robust}]{%
         \includegraphics[height=0.33\textwidth]{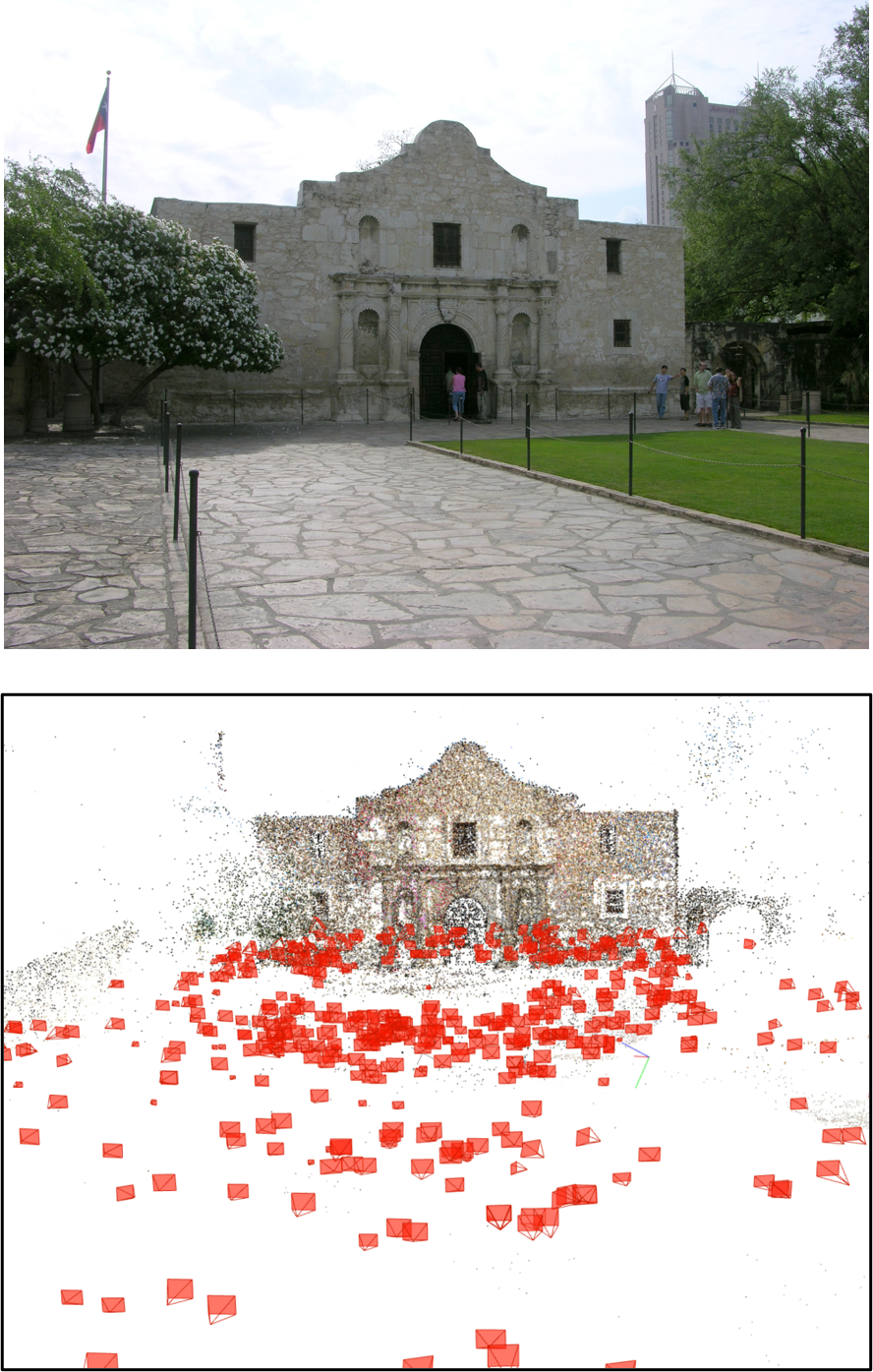}
        }
         \label{fig:examplary_1dsfm}
    \caption{\textbf{Datasets.} 
    Reconstruction examples from EuRoC~\cite{burri2016euroc}, LaMAR~\cite{sarlin2022lamar}, and 1DSfM~\cite{wilson2014robust}; and trajectories from KITTI~\cite{geiger2013vision}. Our method consistently outperforms baselines across these diverse datasets, showcasing its generalization ability.
    }
    \label{fig:examplary}
\end{figure*}

%% file: tex/text/related_work.tex
\section{Related Work}

\noindent\textbf{Global Structure-from-Motion} reconstructs scenes by aggregating pair-wise relative pose estimates. 
This is mostly done via rotation averaging~\cite{olsson2011stable,moulon2013global}.
After determining rotations, translations and structures are recovered.
For instance, Wilson and Snavely~\cite{wilson2014robust} use an outlier filter based on projecting the translations into 1D subspaces.
Moulon et al.~\cite{moulon2013global} employ $L_\infty$-optimization.
Olsson and Enqvist~\cite{olsson2011stable} use $L_\infty$-optimization to solve for 3D points and camera positions.

There are several open-source libraries for global Structure-from-Motion, such as Theia~\cite{theia-manual} and OpenMVG~\cite{moulon2016openmvg}. 
We have incorporated our enhanced rotation averaging method into the Theia framework, keeping the rest of the pipeline unchanged.
Nonetheless, our advancements are versatile and not restricted to this particular pipeline.

\noindent\textbf{Rotation Averaging} has a long history in computer vision.
Govindu~\cite{govindu2001combining} linearizes the problem using quaternions, while Martinec and Pajdla~\cite{martinec2007robust} simplify it by omitting certain non-linear constraints on rotation matrices. 
Wilson et al.~\cite{wilson2016rotations} examine the conditions that make the problem tractable.
Eriksson et al.~\cite{eriksson2018rotation} solve it via strong duality.
Semidefinite programming-based (SDP) relaxation approaches~\cite{arie2012global, fredriksson2012simultaneous} provide optimality guarantees by minimizing the chordal distance~\cite{hartley2013rotation}.
Dellaert et al.~\cite{dellaert2020shonan} sequentially lift the problem into higher-dimensional rotations $\text{SO}(n)$ to avoid local minima where standard numerical optimization techniques may fail~\cite{levenberg1944method,marquardt1963algorithm}.
To handle outliers, various robust loss functions have been investigated \cite{hartley2011l1,chatterjee2013efficient,chatterjee2017robust,sidhartha2021all,zhang2023revisiting}.

Recently, learning-based approaches have emerged. 
NeuRoRa \cite{purkait2020neurora}, MSP \cite{yang2021end}, PoGO-Net \cite{li2021pogo} harness Graph Neural Networks to eliminate outliers and regress absolute camera poses. 
DMF-synch \cite{tejus2023rotation} employs a matrix factorization technique for pose extraction. 
While the above methods offer promising results and runtimes, they operate with 3-DoF rotations, neglecting the advantages of gravity.

\noindent\textbf{Leveraging Gravity Direction} 
has found its way into numerous applications in computer vision and robotics~\cite{petovello2003real,saurer2015homography,liu2019edge,chang2020gnss,feng2021comprehensive}.
Here, we focus on gravity provided by recent consumer devices \textit{by default} through their built-in inertial-based sensory systems \cite{madgwick2011estimation}.
Fraundorfer et al.~\cite{fraundorfer2010minimal} showcase how gravity helps determine yaw and roll in camera rotation, simplifying it to a single degree of freedom.
This simplification extends to rotation averaging, aligning it with planar pose graph optimization techniques.

Planar pose graph optimization was pioneered by Carlone et al.~\cite{carlone2012linear}, designing the LAGO pipeline~\cite{carlone2014fast}. 
While being efficient, LAGO is not robust.\footnote{This will be discussed in depth later in Section~\ref{sec:rot_avg_known_dir}, \ref{sec:circ_reg}.}
Subsequent methods proposed different representations: solving in the complex domain~\cite{carlone2016planar, fan2019efficient} and using SDP relaxation~\cite{carlone2018convex}.
Despite their theoretical strengths, they are not robust to outliers in practical applications.
Using gravity direction in global SfM has barely been explored. 
It has predominantly been viewed as an auxiliary constraint.
Instead of simplifying the problem, it adds complexity to the employed non-linear optimization strategy in prior work.
Both \cite{crandall2012sfm} and \cite{carlone2015initialization} incorporate losses to penalize deviations of the estimated rotational axis from the gravity. 
Experimentally, we will demonstrate that directly optimizing 1-DoF rotation can enhance performance and efficiency.

%% file: tex/text/method.tex
\section{Gravity-Aligned Rotation Averaging}

\noindent\textbf{Rotation Averaging} is a widely studied problem with several improvements proposed throughout the years.
Essentially, the task is to estimate a set of $n \in \mathbb{N}$ rotation matrices $\{\bm{R}_i\} \subset \text{SO}(3)$, given noisy observations of relative rotations $\{\tilde{\bm{R}}_{ij}\} \subset \text{SO}(3)$ estimated in image pairs.
The cameras and the relative rotations form a pose graph with the cameras being the vertices $\mathcal{V}$ and relative poses being the edges $\mathcal{E}$.
The relationship between the absolute $\bm{R}_i$ and relative $\bm{R}_{ij}$ rotations, for all $(i, j)\in \mathcal{E}$, is as follows: 
\begin{equation}
    \bm{R}_{ij} = \bm{R}_j\bm{R}_i^\top,
    \label{eq:r_ij}
\end{equation}
where $\bm{R}_{ij}$ represents the noise-free relative rotation between images. 
In practice, \eqref{eq:r_ij} never holds due to the noise in the estimations. 
Therefore, the problem is typically formulated as a robust least-metric problem as follows:
\begin{equation}
    \arg\min_{\{\bm{R}_i\}} \sum_{(i,j)\in \mathcal{E}} \rho\left(d(\bm{R}_j^\top \tilde{\bm{R}}_{ij} \bm{R}_i, \bm{I})^p\right),
    \label{eq:R_err_3dof}
\end{equation}
where exponent $p \geq 1$, 
$d(\cdot,\cdot)$ is a distance function and 
$\rho(\cdot)$ is a robust loss. 
A comprehensive discussion on potential distance functions is provided by Hartley \cite{hartley2013rotation}. 
In the proposed method, we use geodesic error as follows:
\begin{equation}
    d_{\text{geod}}(\bm{R}_j^\top \tilde{\bm{R}}_{ij} \bm{R}_i, \bm{I}) = \arccos\left(\frac{tr(\bm{R}_j^\top \tilde{\bm{R}}_{ij} \bm{R}_i) - 1}{2}\right),
    \label{eq:r_geod}
\end{equation}
which leads to a simple formulation. 
However, for gradient-based optimization, the chordal distance, defined as the Frobenius norm of the matrix difference 
\begin{equation}
    d_{\text{chor}}(\bm{R}_j^\top \tilde{\bm{R}}_{ij} \bm{R}_i, \bm{I}) = \|\bm{R}_j^\top \tilde{\bm{R}}_{ij} \bm{R}_i - \bm{I}\|_F,
    \label{eq:r_chor}
\end{equation}
can also be used as a differentiable objective function.


\subsection{Rotation Averaging with Known Direction}
\label{sec:rot_avg_known_dir}

The literature on rotation averaging generally considers the 3-DoF problem, where the cameras are rotated arbitrarily in 3D.
However, recent image-capturing devices are equipped with gyroscopes that provide accurate estimates of the gravity direction in the images.
Such directions can also be extracted from vertical vanishing points \cite{lopez2019deep}, which can also be used in this work.
Having such known directions simplifies the problem, as will be shown in this section. 

Let us assume that we are given known direction $\bm{g}_i$ in image $I_i$, which allows for pre-aligning the camera coordinate system with the $y$-axis without loss of generality.  
This pre-alignment is done by rotation $\bm{U}_i$ (s.t.\ $\bm{U}_i\cdot [0, 1, 0]^\top \parallel \bm{g}_i$), reducing the degree-of-freedom by 2 and simplifying the camera rotation as follows: 
\begin{equation}
\bm{R}_i = \bm{U}_i
    \begin{bmatrix}
        \cos \theta_i & 0 & -\sin \theta_i \\
        0 & 1 & 0 \\
        \sin \theta_i & 0 & \cos \theta_i 
    \end{bmatrix}
     = \bm{U}_i \bm{R}_{\theta_i},
\label{eq:r_1dof}
\end{equation}
where $\theta_i \in [-\pi, \pi)$ is an unknown angle around the $y$-axis.
Let us denote the aligned rotations as $\hat{\bm{R}}_i = \bm{U}_i^\top\bm{R}_i= \bm{R}_{\theta_i}$. 
Then \eqref{eq:r_ij} can be reformulated as follows:
\begin{align}
    \bm{R}_{ij} 
    = \bm{U}_j\hat{\bm{R}}_j\hat{\bm{R}}_i^\top\bm{U}_i^\top 
    = \bm{U}_j\begin{bmatrix}
        \cos (\theta_j - \theta_i) & 0 & -\sin (\theta_j - \theta_i) \\
        0 & 1 & 0 \\
        \sin (\theta_j - \theta_i) & 0 & \cos (\theta_j - \theta_i)
    \end{bmatrix}
    \bm{U}_i^\top.
\label{eq:R_ij_1dof}
\end{align}
By setting $\hat{\bm{R}}_{ij} = \bm{U}_j^\top\bm{R}_{ij}\bm{U}_i$, we have $\hat{\bm{R}}_{ij} = \bm{R}_{\theta_{ij}}$, 
%
%
where $\theta_{ij} = \theta_j - \theta_i$. 
Substituting this formula into \eqref{eq:r_ij}, we get
\begin{equation}
    d_{\text{geod}}(\bm{R}_j^\top \tilde{\bm{R}}_{ij} \bm{R}_i, \bm{I})
    = \arccos(\cos[\tilde{\theta}_{ij} - (\theta_j - \theta_i)]) \label{eq:theta_ij} 
    = [\tilde{\theta}_{ij} - (\theta_j - \theta_i)] + 2k_{ij}\pi,
    \nonumber
\end{equation}
where $k_{ij} \in \mathbb{Z}$. 
By plugging this into the robust least-metric problem, \eqref{eq:R_err_3dof} becomes
\begin{equation}
    \arg \min_{\{\theta_i\},\{k_{ij}\}\subset\mathbb{Z}} \sum_{(i,j)\in \mathcal{E}} \rho\left(d(\tilde{\theta}_{ij} - (\theta_j - \theta_i) + 2k_{ij}\pi, 0)^p\right).
\label{eq:R_err_1dof_k}
\end{equation}
%
Note that this formula is non-linear due to the periodicity of variables $\theta_i$, so standard linear solvers fail.
LAGO \cite{carlone2014fast} addresses this by pre-estimating the periods from a coarse initialization obtained by analyzing the camera trajectory and keeping them fixed afterwards.
Such a strategy fails under erroneous initialization, especially in the SfM case that lacks continuous trajectories. 

\input{tex/fig/result_full}

\subsection{Circular Regression}

Circular regression models are proposed to solve systems that contain periodic variables.
Interestingly, this periodic nature of data and its implications have been widely recognized in the field of statistics but not in computer vision. Here, we briefly introduce the problem and refer interested readers to \cite{fisher1992regression} for a more comprehensive discussion.   

In general, circular regression falls in the study of circular analysis, which aims to tackle the periodicity of circular data. 
Besides the angles in this work, other forms of circular data also widely exist in nature and are studied across fields such as ecology, medical sciences, and cognitive and experimental psychology~\cite{cremers2018one}. 
Since there is no natural scale measure for circular data, the von Mises distribution (also known as circular normal distribution) is the alternative for the standard Gaussian distribution~\cite{mardia1975statistics}. 
%
Density function $f$ of a von Mises $\text{VM}(\mu, \kappa)$ distribution is of the following form
\begin{equation}
    f(\theta; \mu, \kappa) = [2\pi I_0(\kappa)]^{-1} \exp[\kappa \cos(\theta-\mu)], 
\end{equation}
where $I_0(\kappa)$ is a scaling factor such that 
%
    $\int_{-\pi}^\pi \exp (\kappa\cos x) dx = 2\pi I_0(\kappa).$
%
Parameter $\mu$ is the cluster centers of the distribution while $\kappa$ is a reciprocal measure of dispersion ($\kappa\approx 1/\sigma^2$).
It is the circular analog of the normal distribution wrapped around a specific range.
When $\kappa \geq 2$, density at $\mu + \pi$ is effectively zero, and it can be well-approximated by a normal distribution. 

Let us consider the problem at hand and assume that each noisy rotation $\tilde{\theta}_{ij}$ is drawn independently and follows a von Mises distribution such that
%
    $\tilde{\theta}_{ij} \sim f(\theta; \theta_j - \theta_i, \kappa_{ij})$.
%
The probability of the current set of observations $\{\tilde{\theta}_{ij}\}$ is
\begin{equation}
    \text{Pr}(\{\tilde{\theta}_{ij}\}) = \prod_{(i,j)\in \mathcal{E}} f(\tilde{\theta}_{ij}; \theta_j - \theta_i, \kappa_{ij}).
    \label{eq:vm_mle}
\end{equation}
The maximum likelihood estimation of \eqref{eq:vm_mle} can be approximated by the following form as in normal distribution:
\begin{equation}
    \min_{\{\theta_i\}} \sum_{(i,j)\in \mathcal{E}} \kappa_{ij}\left(\tilde{\theta}_{ij} - (\theta_j - \theta_i) + 2k_{ij}\pi\right)^2,
\label{eq:R_err_mle}
\end{equation}
where $[\tilde{\theta}_{ij} - (\theta_j - \theta_i) + 2k_{ij}\pi] \in [-\pi,\pi)$. 
We assume $\kappa_{ij}$ to be the same for all relative rotations, but they can be leveraged to formulate uncertainty-aware rotation averaging as in \cite{zhang2023revisiting}.
Notice that \eqref{eq:R_err_1dof_k} is a generalized form of the equation, validating our design choice. 
Let us note that the proposed formulation follows a similar principle as in Barber's pole model proposed by Gould~\cite{gould1969regression}, and we can similarly apply an iterative procedure to solve the problem.
%

\input{tex/algorithm_circreg}
\label{sec:circ_reg}
The standard procedure minimizing \eqref{eq:R_err_1dof_k} is an iterative optimization.
In each step, the algorithm first decides which period angular variable $\tilde{\theta}_{ij}$ fits in by finding $k_{ij}$ for each term. 
Then it solves a linear regression problem, solving linear system \eqref{eq:theta_ij}, with period $k_{ij}$ being fixed.
It iterates until convergence, as measured by the change in the estimated variables. Let us denote 
\begin{align}
     k_{\theta_{i}, \theta_{j}}^* &= \argmin_{k_{ij}\in\mathbb{Z}} |\tilde{\theta}_{ij} - (\theta_j - \theta_i) + 2k_{ij}\pi|, \\
     \hat{\epsilon}_{ij} &= \tilde{\theta}_{ij} - (\theta_j - \theta_i) + 2k_{ij}\pi, \quad\hat{\epsilon}_{ij}\in\mathbb{R}.
\label{eq:k_star}
\end{align}
The process is described in Alg.~\ref{alg:opt_err_1dof}.
During the iterations, the error in period estimation is gradually rectified.
In contrast, the error in LAGO-like methods~\cite{carlone2014angular,fan2019efficient} persists, resulting in a large performance gap.
See Supp.\ Mat.\ for details.
Note that the algorithm is not guaranteed to converge to global optimal.
However, in practice, we have not found a case where the algorithm stuck at a local minimum with a large error. 
For solving the global optimum, the problem can also be formulated as a Mixed Integer Programming, which is NP-Hard \cite{wolsey1999integer} and thus impractical for real-world and large-scale problems.


\noindent\textbf{Convergence Properties.}
Denote $d_\text{geod}(\bm{R}_j^\top \tilde{\bm{R}}_{ij} \bm{R}_i, \bm{I})$ as $\bar{\epsilon}_{ij}$, then
by enforcing $\tilde{\theta}_{ij}, \theta_i, \theta_j \in [-\pi, \pi)$, the residual after each iteration is formalized as
\begin{equation}
    \epsilon_{ij} = \tilde{\theta}_{ij} - (\theta_j - \theta_i) = \bar{\epsilon}_{ij} + 2k_{\theta_i \theta_j}^*\pi ,
    \label{eq:residual}
\end{equation}
where the number of choices for $k_{\theta_{i}, \theta_{j}}^* \in \{-1, 0, 1\}$ is finite. 
We remove the gauge freedom by setting $\theta_1 = 0$, and choose $d(\cdot)$ to be convex. 
By this, we circumvent the problem of the original Barber pole's model, where we are given an infinite number of peaks (\ie, optimal solutions) with the same probability.
 
The proposed algorithm inherits the convergence properties of the underlying linear solver. 
In each iteration, $k_{ij}$ are set to minimize the respective error.
Thus, the change of $k_{ij}$ is guaranteed not to increase $d(\hat{\epsilon}_{ij})$. 
Since this holds for all terms, the total error cannot increase. 
This means that the overall error after each iteration is monotonically decreasing. Otherwise, the algorithm terminates.
Notice that the total number of configurations for $k_{ij}$ is finite, and a configuration cannot be encountered twice due to the strict monotonicity. 
Thus, the algorithm will always terminate. 
Consequently, the convergence solely depends on the underlying linear solver which has its own strong convergence guarantees. 


\subsection{Partially Known Gravity}

The assumption that gravity direction is available in all images may be invalid when images are collected from various sources. 
Thus, we extend our model for when a subset of images has no known gravity.
Chatterjee et al.~\cite{chatterjee2013efficient} propose an efficient rotation averaging approach based on the axis-angle representation, minimizing the geodesic error.
The distance between two rotations represented by axis-angle $\theta\cdot\bm{v} \in \mathfrak{so}(3)$ approximates the geodesic distance, enabling the linearization of the relative rotation constraint.

We utilize the angle-axis representation for images without known gravity while preserving the single-variable form for those with an established gravity direction. 
This categorization results in three types of relative pose constraints: 
both images have known gravity; 
neither image has gravity; 
or one image has gravity while the other lacks it. 
The proposed method directly addresses the first scenario. 
The second scenario is handled based on \cite{chatterjee2013efficient}. 
For the third scenario, we employ a linear system comprised of three constraints, in which the first and third components of $\theta\cdot\bm{v}$ are fixed at zero for images with gravity direction.
All equations are collected together to form a mixed linear system.
We take a stratified approach to solve it.
We first extract the largest connected component for images with gravity and perform the 1-DoF rotation averaging on images within this component.
Afterward, we set the remaining rotation matrices to be identity matrices and solve the mixed linear system. 
Such a design can leverage the fast convergence of 1-DoF images and works well in our experiments.

\subsection{Gravity Direction Refinement}
\label{sec:refinement}

Here, we provide a heuristic approach to improve gravity directions.
In many cases, the inaccuracy of the gravity of a particular image can be recognized by analyzing the set of estimated 3-DoF relative poses between the image and its neighbors. 
From Eq.~\eqref{eq:R_ij_1dof}, noisy relative rotation can be pre-aligned as
\begin{equation}
    \tilde{\bm{R}}_{\theta_{ij}} = \bm{U}_j^\top \tilde{\bm{R}}_{ij} \bm{U}_i.
\label{eq:align_Rij}
\end{equation}
Rotations $\tilde{\bm{R}}_{\theta_{ij}}$ provide signal on the validity of $\bm{U}_i$ and $\bm{U}_j$.
Ideally, $\tilde{\bm{R}}_{\theta_{ij}}$ should be a rotation with only 1 DoF.
If the yaw and roll of $\tilde{\bm{R}}_{\theta_{ij}}$ deviate from 0, there are two possibilities: the relative rotation is inaccurate, or at least one of the two gravities is inaccurate. 
Based on such an observation, we propose a majority voting mechanism to identify error-prone gravity directions.
For each image, we maintain a counter.
When $\tilde{\bm{R}}_{\theta_{ij}}$ has a large yaw or roll, we increment the counter for both images. 
If the counter for a image is larger than half of its neighbor number, we refine the gravity direction. 

Eq.~\eqref{eq:align_Rij} also allows for refining gravity direction $\bm{g}_{j}$ of the $j$-th image. 
To estimate $\bm{g}_j$ given $\tilde{\mathbf{R}}_{ij}$ and $\mathbf{U}_i$, we have
\begin{equation}
    \bm{U}_j  \tilde{\bm{R}}_{\theta_{ij}} = \tilde{\bm{R}}_{ij} \bm{U}_i 
    \Rightarrow 
    \bm{g}_{j} = (\bm{U}_j)_{:,1} = (\bm{U}_j \tilde{\bm{R}}_{\theta_{ij}})_{:,1} = (\tilde{\bm{R}}_{ij} \bm{U}_i)_{:,1},
\end{equation}
where subscript $(\bm{U}_j)_{:,1}$ selects the second column (\ie, down direction) of $\bm{U}_j$.
Given multiple neighbors of view $i$, we perform robust estimation of the directions.
We minimize the sum of differences between the estimation and results from neighbors.
To achieve robustness, we optimize the Arctan robust loss using Ceres \cite{ceres_solver}.
%
Sec.~\ref{sec:ablation_refine} demonstrates the effectiveness of this refinement strategy.





%% file: tex/fig/result_full.tex
\begin{figure*}[t]
    \centering
     \includegraphics[width=0.9\textwidth]{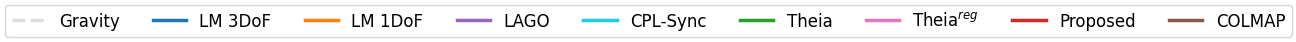}\\
     \begin{subfigure}[b]{0.255\textwidth}
         \centering
         \includegraphics[width=\textwidth]{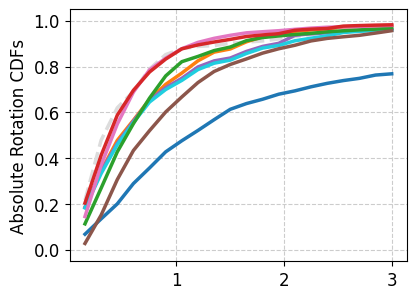}
         \caption{EuRoC \cite{burri2016euroc}}
         \label{fig:result_euroc}
     \end{subfigure}
     \begin{subfigure}[b]{0.24\textwidth}
         \centering
         \includegraphics[width=\textwidth]{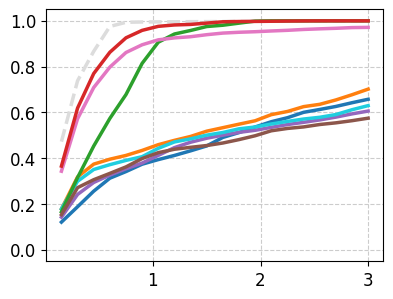}
         \caption{KITTI \cite{geiger2013vision}}
         \label{fig:result_kitti}
     \end{subfigure}
     \begin{subfigure}[b]{0.24\textwidth}
         \centering
         \includegraphics[width=\textwidth]{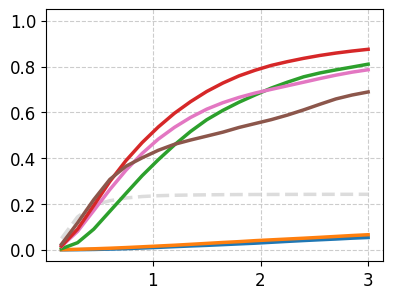}
         \caption{LaMAR \cite{sarlin2022lamar}}
         \label{fig:result_lamar}
     \end{subfigure}
     \begin{subfigure}[b]{0.24\textwidth}
         \centering
         \includegraphics[width=\textwidth]{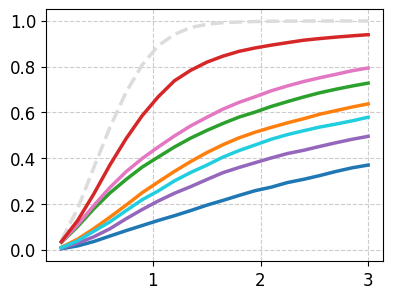}
         \caption{1DSfM \cite{wilson2014robust}}
         \label{fig:result_1dsfm}
     \end{subfigure}
    \caption{ 
    The cumulative distribution functions (CDFs) of the absolute rotation errors ($^\circ$). Estimated by the Levenberg-Marquardt~\cite{more2006levenberg} method solving the 3-DoF (LM 3DoF) and 1-DoF (LM 1DoF) problems, by LAGO~\cite{carlone2014fast}, by CPL-Sync~\cite{fan2019efficient}, by the rotation averaging in the Theia library~\cite{theia-manual,chatterjee2013efficient}, by Theia with an additional penalty term (Theia$^\text{reg}$ \cite{crandall2012sfm}), by COLMAP~\cite{schoenberger2016sfm}, and by the proposed method.
    Curve "Gravity" stands for the approximate upper bound achievable by using gravity direction.
    A method being accurate is interpreted by its curve close to the top-left corner. 
    Different metrics for the datasets are reported in Tables~\ref{tbl:euroc}, \ref{tbl:KITTI}, \ref{tbl:lamar_map}, \ref{tbl:1dsfm}. }
    \label{fig:result_full}
\end{figure*}

%% file: tex/algorithm_circreg.tex

\begin{figure}[t]
\begin{center}
    \scalebox{0.85}{
\begin{minipage}{\textwidth}
      \vspace{0pt}
\begin{algorithm}[H] \small
\caption{Solve optimization for \eqref{eq:R_err_1dof_k}}
\label{alg:opt_err_1dof}
\DontPrintSemicolon
\SetAlgoLined
\SetNoFillComment
$\theta_i\gets 0$ if $\theta$ \text{not initialized}\;
$k_{ij} \gets k_{\theta_{i}, \theta_{j}}^*$ \;
\While{not converged \textbf{and} $ite < max\_ites $} {
    $\theta_i \gets \argmin_{\{\theta_i\} } \sum_{(i,j)\in \mathcal{E}} \rho\left(d(\hat{\epsilon}_{ij})^p\right)$ \tcp*{optimize $\theta_i$ with fixed $k_{ij}$}
    $k_{ij} \gets k_{\theta_{i}, \theta_{j}}^*$ \tcp*{optimize $k_{ij}$ with fixed $\theta_i$}
}
\end{algorithm}
\end{minipage}
}
\end{center}
\end{figure}

%% file: tex/text/experiments.tex
\section{Experiments}
%

\noindent\textbf{Implementation.}
We implemented the proposed rotation averaging in C++ in a similar fashion as the widely used Theia library \cite{theia-manual}.
%

\noindent\textbf{Distance Functions and Robust Loss.}
Having an ordinary linear system in each step of the algorithm allows us to use any $n$-norm as a distance function and robust loss, such as Huber \cite{huber1992robust} or Cauchy Loss \cite{black1996robust}.
To be robust to outliers, we follow the two-stage optimization strategy of Theia \cite{theia-manual}.
In the first stage, we minimize $L_1$ distance. 
In the second one, we use an iteratively re-weighted approach.
The weights are determined by the Geman-McClure function \cite{ganan1985bayesian}.
The parameter choice follows the Theia library~\cite{theia-manual}.

\noindent\textbf{Competitors.}
We compare the proposed rotation averaging to
LAGO \cite{carlone2014fast}, implemented in GTSAM library \cite{gtsam}, to CPL-Sync \cite{fan2019efficient},  
with \cite{chatterjee2013efficient} and \cite{crandall2012sfm} implemented in the Theia library~\cite{theia-manual},
and with methods optimizing the 3-DoF and 1-DoF problems by the Levenberg–Marquardt (LM) algorithm \cite{levenberg1944method} with the Ceres library \cite{ceres_solver}.
CPL-Sync~\cite{fan2019efficient} is a planar PGO algorithm.
For the LM optimization, we use the Soft-$L_1$ robust loss function \cite{ceres_solver}.
To be differentiable and exploit the periodicity, we use the scaled $\sin$ as distance function as 
%
    $d_{\text{1-DoF}}(\epsilon) = 2\sqrt{2}\sin (\epsilon/2)$. 
%
Note that the error is equivalent to the chordal error \cite{hartley2013rotation}. 
We use the chordal error in the 3-DoF optimization for a fair comparison.

Additionally, we compare with another baseline.
Similarly as in \cite{crandall2012sfm}, we add an additional error term penalizing the angle between the rotation axis and the gravity direction.
The objective function becomes:
\begin{equation}
    \arg\min_{\{\bm{R}_i\}}E_{\mathcal{E}} + \sum_{i} \lambda \rho\left(d(\theta \cdot \left(\bm{v}_i - (\bm{v}_i^\top\bm{g}_i)\bm{g}_i)\right)^p\right),
    \label{eq:R_err_penalty}
\end{equation}
where $E_{\mathcal{E}}$ is the same as Eq. \eqref{eq:R_err_3dof}, $\theta\cdot\bm{v}_i \equiv \bm{R}_i$ is the axis-angle representation of $\bm{R}_i$, 
$\bm{g}_i$ is the gravity, $d$ is the distance function, $\rho$ is the robust loss and $\lambda$ controls the level of regularization.
In the experiments, we set $\lambda$ to 1 as in \cite{carlone2015initialization} and denote this method as \textit{Theia$^\text{reg}$}.
See implementation details in supp. mat.
In practice, it is possible to modify the Theia~\cite{theia-manual} pipeline to only update the rotation around $y$-axis.
While the objective function of the proposed method and a modified Theia formulation are the same, the latter solves it through a more complex procedure.
As a result, the running time of such a formulation is several times slower than the proposed method.
Also, it is not applicable when the gravity is partially known.
See details in supp. mat.

We also tested recent learning-based alternatives \cite{purkait2020neurora, yang2021end, li2021pogo, tejus2023rotation}.
While they report promising results on the 1DSfM datasets~\cite{wilson2014robust}, they fall short on the tested sequential ones \cite{burri2016euroc, geiger2013vision, sarlin2022lamar}. 
This is likely caused by the connectivity of the pose graph being too sparse.
We included their results in the Supplementary Material.

\input{tex/tbl/euroc}

While we compare to global SfM algorithms, we also show the results of COLMAP \cite{schoenberger2016sfm} (\ie, incremental SfM).
Even though COLMAP produces high-quality reconstructions, it runs significantly slower than global SfM in practice.

\noindent\textbf{Comparison.}
In all tests, the relative poses are estimated by LO-RANSAC \cite{chum2003locally}, obtaining the 5-DoF essential matrix \textbf{E}.
Experiments with 3-DoF \textbf{E} are included in the supp. mat.
In short, it improves the results for both our method and Theia~\cite{theia-manual} but the relative performance remains the same.
To address the gauge ambiguity in comparing with the ground truth (GT), we perform a global 3-DoF rotation alignment by minimizing the Cauchy Loss using Ceres~\cite{ceres_solver}.
Geodesic error $d_{\text{geod}}$ between estimated $\bm{R}_i$ and GT $\bm{R}^{gt}_i$ is considered as rotation error.

\subsection{Rotation Averaging with Measured Gravity}

\noindent\textbf{EuRoC \cite{burri2016euroc}} is a visual-inertial dataset collected by a Micro Aerial Vehicle (Figure \ref{fig:examplary_euroc}).
The dataset contains stereo images and synchronized IMU measurements and is millimeter-accurate.
We align the gravity to the first frame and use \cite{madgwick2010efficient} from \cite{madgwick_filter} to calculate gravity direction from the raw measurements.
On average, the sequences comprise 1077 images and 12620 pairwise motion estimates.

The average and median rotation errors in degrees, the Area Under the recall Curve (AUC) thresholded at $0.5^\circ$, $1^\circ$, and $2^\circ$, and the average runtime (in hours) are reported in Table \ref{tbl:euroc}.
The 1-DoF optimization performs significantly more accurately and faster than the 3-DoF optimization, especially when the gravity is accurate, motivating its use. 
Also, the method implemented in the Theia library is significantly more advanced than a simple LM optimization, achieving better accuracy than the 3-DoF and 1-DoF optimizations. 
Compared to LAGO and CPL-Sync, the proposed method achieves much higher accuracy and is more efficient.
The same holds compared to Theia \cite{theia-manual}.
When compared to Theia with an extra penalty term (Theia$^\text{reg}$), we achieve better or comparable results in all metrics while being an order of magnitude faster. 
The proposed method is substantially faster and more accurate than COLMAP in all metrics. 

The cumulative distribution functions (CDF) of the absolute rotation errors are in Fig. \ref{fig:result_euroc}. 
A method being accurate is interpreted as its curve close to the top-left corner.
Curve ``Gravity'' (dashed curve) shows the accuracy of gravity bound, and is in principle the upper bound of the gravity-based methods. 
It is calculated as the geodesic error between the GT rotation and the closest rotation matrix, which aligns with the current gravity direction. 
Precisely, the error is 
%
    $\epsilon_{\text{grav}, i} = d_{\text{geod}}(\bm{R}_i^\top\bm{\tilde{U}}_{i}\bm{R}_{\theta_{i}}, \bm{I})$,
%
where $\bm{\tilde{U}}_i$ aligns the noisy gravity. 
Our method closely approaches the gravity upper bound curve.

\input{tex/tbl/KITTI}

\noindent\textbf{KITTI \cite{geiger2013vision}} is a real-world autonomoust driving visual odometry benchmark with GT from GPS localization system (Fig.~\ref{fig:examplary_kitti})
%
We tested our method on the 9 odometry sequences (without 00, 03) provided with ground truth. 
For sequence 00, several data intervals are missing, and there is no raw IMU data for sequence 03.
In the tested sequences, there are 1984 images and we form 20767 image pairs on average.
We use the Invariant Extended Kalman Filter implemented in \cite{brossard2020ai} to estimate the gravity direction from raw synchronized IMU data.

The results are in Table \ref{tbl:KITTI}.
The proposed method outperforms Theia~\cite{theia-manual} by a large margin, featuring 23, 22, and 12 points improvement in AUC score at $0.5^\circ$, $1^\circ$ and $2^\circ$ respectively, and is about 8 times faster.
Compared to Theia$^\text{reg}$ \cite{crandall2012sfm}, the proposed method achieves substantial accuracy improvements while running 30 times faster.
Even though LAGO and CPL-Sync are designed for this scenario, they perform inaccurately due to the missing of a robust objective function.
Compared to COLMAP, the estimation is significantly more accurate and efficient.
Due to the estimation drift, COLMAP~\cite{schoenberger2016sfm} accumulates large errors along the sequence, while the global methods generally achieve low estimation error. 

The cumulative distribution functions (CDF) of the absolute rotation errors are shown in Fig.~\ref{fig:result_kitti}. 
The proposed method (red) is the closest to the top-left corner.
It approaches the gravity upper bound closely and is significantly more accurate than the other compared algorithms.  
%

\input{tex/tbl/slam_compare}
Table \ref{tbl:slam_compare} compares the proposed method with SLAM systems.
The SLAM results (provided on 5 sequences) are copied from \cite{sharafutdinov2023comparison}. 
The proposed algorithm is the most accurate on EuRoC by a large margin.
It is the best on two sequences of KITTI, and, on the third, it is on par with the best one with only $0.12^\circ$ difference. 
Note that several SLAM pipelines also take IMU data into consideration and, potentially, other sensors, such as VINS-Fusion~\cite{qin2019general} that uses stereo cameras. 
These results clearly demonstrate the robustness of the proposed method. 

\subsection{Experiments with Partially Known Gravity}

\noindent\textbf{LaMAR \cite{sarlin2022lamar}} is a recent large-scale benchmark. 
Example images and reconstruction can be found in Figure \ref{fig:examplary_lamar}.
The dataset contains images captured by Hololens 2 \cite{hololens} and smartphones.
Smartphone images are processed with ARKit~\cite{arkit}, estimating gravity-aligned trajectory via onboard inertial units. 
Such estimations are available \textit{by default}, and can be readily used in applications.
We use gravity information captured by smartphones and all sequences from three scenes.
The considered sequences consist of 31676 images with 194182 image pairs on average, and 25\% of images are with gravity directions.
Since we focus on the partial gravity case here, we do not consider the gravity from the Hololens devices. 
Additional results are in the Supplementary Material. 

%
\input{tex/tbl/lamar_map}
Results are in Table \ref{tbl:lamar_map}.
We do not show results for LAGO~\cite{carlone2014angular} and CPL-Sync~\cite{carlone2014fast} as they are not applicable in this scenario.
The proposed method outperforms Theia-based methods in all metrics while running 1.5 times faster.
It is on par with COLMAP, having a marginally lower AUC $0.5^\circ$ score and being better in other metrics.
However, COLMAP runs four orders of magnitude slower, in a matter of hours instead of seconds.

The cumulative distribution functions (CDF) of the absolute rotation errors are shown in Fig. \ref{fig:result_lamar}. 
The proposed method (red) is the closest to the top-left corner, showing that it leads to the lowest absolute errors. 

\subsection{Semi-Synthetic Experiments}


\noindent\textbf{1DSfM \cite{wilson2014robust}} contains Internet photo collections, featuring the Structure-from-Motion task on unordered image sets. 
The gravity direction for each image is not provided. Thus, we use the ground truth direction and add zero-mean Gaussian noise. 
We set the standard deviation of the gravity error to $0.5^\circ$. 
This is slightly larger than the noise levels observed in the EuRoC, KITTI, and LaMAR datasets. 
We consider the reconstruction in the original dataset as ground truth and use the relative poses provided as input.

\input{tex/tbl/1dsfm}
The results are summarized in Table \ref{tbl:1dsfm}. 
For all methods, we used improved gravities as input.
We see similar trends to those of the other datasets. 
The direct 1-DoF optimization improves significantly upon the 3-DoF method. 
The proposed algorithm leads to the lowest median error and the highest AUC scores.
It is about 15 times faster than the Theia-based methods.
Compared with LAGO~\cite{carlone2014fast} and CPL-Sync~\cite{fan2019efficient}, the proposed method obtains significantly more accurate results. 
%
For this dataset, we do not report COLMAP results as the ground truth is from another structure-from-motion method. Thus, the error is not directly comparable.
The cumulative distribution functions (CDF) of the absolute rotation errors are shown in Fig. \ref{fig:result_1dsfm}.
The proposed method (red) is the closest to the top-left corner, and it clearly outperforms other baselines.

\noindent
\textbf{Ablation Study for Gravity Refinement.\label{sec:ablation_refine}}
Table \ref{tbl:refinement} shows the refinement performance on the 1DSfM~\cite{wilson2014robust} dataset.
Even with noisy relative pose observations, rotation averaging largely benefits from performing gravity direction refinement.
The improvement is especially significant for the AUC scores at $0.5^\circ$ and $1^\circ$.
The runtime overhead is less than $10\%$. 
Note that we observed no improvements on the sequence-based datasets. 
We attribute this to the fact that the gravity noise is not independent in such cases, \eg, it stems from drifting. 

\input{tex/tbl/refine}

\subsection{Synthetic Experiments}

To show the scalability of the gravity-aligned optimization, we conduct tests by generating graphs with a varying number of edges and cameras.
We experiment with two types of view graphs, \ie, sequentially and grid-connected ones, to mimic the image stream and unordered image sets settings. 
For the sequential setting, each camera is connected to its 20 nearest neighbors, which leads to graphs with similar connectivity as the KITTI and EuRoC datasets. 
For grid data, each camera is connected with its 24 closest neighbors to have a similar number of neighbors to the sequential case. 

\begin{figure}[t]
    \centering
    \includegraphics[width=0.7\columnwidth]{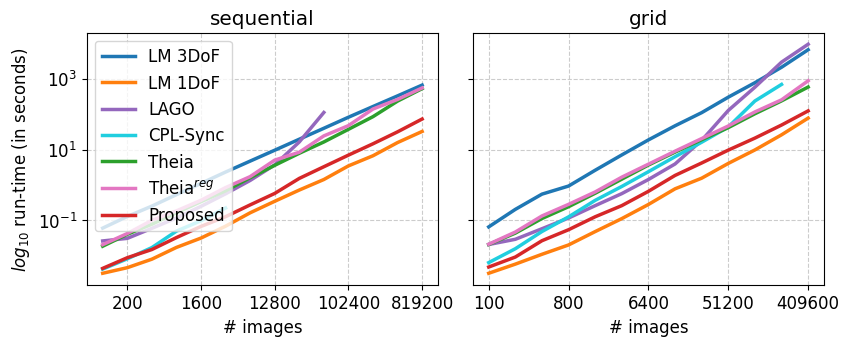}
    \caption{
    Avg.\ $\log_{10}$ time (secs) as a function of image number in synthetic pose graphs: sequential (left) and grid-like (right) cameras.
    }
    \label{fig:runtime}
\end{figure}

The results are summarized in Figure \ref{fig:runtime}. 
For all methods except for LAGO and CPL-Sync, the runtime depends linearly on the number of images in the graph.
Note that the edge number is a constant times the image number. 
The runtime of LAGO undergoes a sudden increase after the images exceed 25600, and the implementation \cite{gtsam} crashes unexpectedly when the number of cameras reaches 102400 for the sequential case.
The proposed method is an order of magnitude faster than Theia-based baselines and the 3-DoF optimization.
%
The reason is two-fold: we achieve a faster convergence while solving a problem with fewer parameters. 
Both lead to a more efficient and stable algorithm. 
The proposed method is several times faster than LAGO~\cite{carlone2014angular}, mainly due to pipeline design.




%% file: tex/tbl/euroc.tex
\begin{table}[t]
    \centering
    \caption{Average and median rotation errors ($^\circ$), the AUC score at $0.5^\circ$, $1^\circ$, and $2^\circ$ and average run-time (hour) on EuRoC~\cite{burri2016euroc}.}
    \resizebox{0.75\columnwidth}{!}{
    \begin{tabular}{l c c c c c c} \toprule 
        & AVG ($^\circ$) & MED ($^\circ$) & AUC@$0.5^\circ$ &~~~~@$1^\circ~~~~$ &~~~~@$2^\circ~~~~$  & Time (h) \\ \midrule
        LM 3DoF & 3.93 & 0.96 & 11.13 & 23.38 & 41.33 & 2.94 $\times 10^{-4}$ \\
        LM 1DoF & 0.74 & 0.43 & 27.55 & 46.02 & 66.47 & \textbf{1.11 $\times 10^{-5}$} \\
        LAGO \cite{carlone2014fast} & 0.83 & 0.45 & 27.18 & 45.41 & 64.41 & 8.61 $\times 10^{-5}$ \\
        CPL-Sync \cite{fan2019efficient} & 0.82 & 0.45 & 26.75 & 44.82 & 63.64 & 2.67 $\times 10^{-4}$ \\
        Theia \cite{theia-manual} & 0.82 & 0.53 & 22.14 & 43.85 & 66.06 & 4.25 $\times 10^{-4}$ \\
        Theia$^\text{reg}$ \cite{crandall2012sfm} & \textbf{0.63} & 0.40 & 29.20 & 52.86 & 72.72 & 5.83 $\times 10^{-4}$ \\
        Proposed & 0.64 & \textbf{0.35} & \textbf{33.06} & \textbf{54.81} & \textbf{72.97} & 5.56 $\times 10^{-5}$ \\
        \hdashline
        COLMAP \cite{schoenberger2016sfm} & 1.17 & 0.77 & 12.93 & 32.27 & 55.82 & 5.05 \\

      \bottomrule
    \end{tabular}}
    \label{tbl:euroc}
\end{table}

%% file: tex/tbl/KITTI.tex
\begin{table}[t]
    \centering
    \caption{Average and median rotation errors ($^\circ$), the AUC score at $0.5^\circ$, $1^\circ$, and $2^\circ$ and average run-time (hour) on KITTI~\cite{geiger2013vision}.
    }
    \resizebox{0.75\columnwidth}{!}{
    \begin{tabular}{l c c c c c c} \toprule 
        & AVG ($^\circ$) & MED ($^\circ$) & AUC@$0.5^\circ$ &~~~~@$1^\circ~~~~$ &~~~~@$2^\circ~~~~$  & Time (h) \\ \midrule
        LM 3DoF & 3.14 & 2.73 & 15.60 & 24.80 & 35.29 & 5.28 $\times 10^{-4}$ \\
        LM 1DoF & 2.77 & 2.36 & 24.42 & 32.84 & 41.88 & \textbf{1.67 $\times 10^{-5}$} \\
        LAGO \cite{carlone2014fast} & 5.50 & 3.17 & 19.14 & 27.15 & 37.38 & 8.06 $\times 10^{-5}$ \\
        CPL-Sync \cite{fan2019efficient} & 4.38 & 2.95 & 23.26 & 31.08 & 40.17 & 5.00 $\times 10^{-5}$ \\
        Theia \cite{theia-manual} & 0.68 & 0.64 & 25.96 & 47.48 & 71.77 & 3.92 $\times 10^{-4}$ \\
        Theia$^\text{reg}$ \cite{crandall2012sfm} & 0.54 & 0.31 & 45.38 & 64.85 & 78.83 & 1.56 $\times 10^{-3}$ \\
        Proposed & \textbf{0.39} & \textbf{0.29} & \textbf{48.71} & \textbf{69.79} & \textbf{83.86} & 5.56 $\times 10^{-5}$ \\
        \hdashline
        COLMAP \cite{schoenberger2016sfm} & 3.76 & 3.13 & 20.09 & 28.29 & 36.91 & 8.80 \\
      \bottomrule
    \end{tabular}}
    \label{tbl:KITTI}
\end{table}

%% file: tex/tbl/slam_compare.tex
\begin{table}[t]
    \centering
    \caption{Comparison with SLAM-based method. Root-mean-square rotation error ($^\circ$) are reported. 
    The SLAM results (only 5 are available) are copied from \cite{sharafutdinov2023comparison}
    }
    \resizebox{\columnwidth}{!}{
    \begin{tabular}{lrccccccc}
    \toprule
    \multicolumn{2}{c}{Dataset} & ORB-SLAM2\cite{mur2017orb} & LDSO\cite{gao2018ldso} & VINS-Fusion\cite{qin2019general} & OpenVSLAM\cite{sumikura2019openvslam} & Basalt\cite{usenko2019visual} & ORB-SLAM3\cite{campos2021orb} & Proposed \\
    \midrule
    \multirow{2}{*}{EuRoC} & V1\_01 & 4.70 & \phantom{1}4.68 & 5.60 & 4.73 & 5.36 & 5.94 & \textbf{2.16} \\
     & MH\_05 & 6.60 & 20.45 & 1.87 & 6.54 & 0.67 & 0.93 & \textbf{0.21} \\
     \cmidrule{1-9}
    \multirow{3}{*}{KITTI} & 02 & 1.47 & \phantom{1}2.41 & 4.78 & 1.20 & 1.72 & 1.43 & \textbf{0.62} \\
    & 05 & 0.38 & \phantom{1}0.87 & 3.02 & 0.38 & 0.86 & \textbf{0.37} & 0.49 \\
    & 06 & 0.39 & \phantom{1}0.59 & 4.81 & 0.69 & 2.78 & 0.41 & \textbf{0.25} \\
    \bottomrule
    \end{tabular}}
    \label{tbl:slam_compare}
\end{table}

%% file: tex/tbl/lamar_map.tex
\begin{table}[t]
    \centering
    \caption{Average and median rotation errors ($^\circ$), AUC score at $0.5^\circ$, $1^\circ$, and $2^\circ$ and average run-time (hours) on LaMAR~\cite{sarlin2022lamar}. 
     }
    \resizebox{0.75\columnwidth}{!}{
    \begin{tabular}{l c c c c c c} \toprule 
        & AVG ($^\circ$) & MED ($^\circ$) & AUC@$0.5^\circ$ &~~~~@$1^\circ~~~~$ &~~~~@$2^\circ~~~~$  & Time (h) \\ \midrule
        LM 3DoF & 51.79 & 35.53 & 0.06 & \phantom{a}0.30 & \phantom{a}1.14 & 4.28 $\times 10^{-2}$ \\
        LM 1DoF & 49.50 & 28.18 & 0.20 & \phantom{a}0.60 & \phantom{a}1.70 & 2.67 $\times 10^{-2}$ \\
        Theia \cite{theia-manual} & \phantom{a}3.46 & \phantom{a}1.28 & 3.28 & 13.92 & 34.62 & 3.72 $\times 10^{-2}$ \\
        Theia$^\text{reg}$ \cite{crandall2012sfm} & 11.77 & \phantom{a}1.03 & 7.09 & 20.69 & 40.40 & 3.68 $\times 10^{-2}$ \\
        Proposed & \phantom{a}\textbf{2.93} & \phantom{a}\textbf{0.93} & 7.94 & \textbf{23.17} & \textbf{45.51} & \textbf{2.41} $\times 10^{-2}$ \\
        \hdashline
        COLMAP \cite{schoenberger2016sfm} & 34.18 & \phantom{a}1.42 & \textbf{9.63} & 22.59 & 36.09 & 98.52 \\
      \bottomrule
    \end{tabular}
    }
    \label{tbl:lamar_map}
\end{table}

%% file: tex/tbl/1dsfm.tex
\begin{table}[t]
    \centering
    \caption{Average and median rot.\ errors ($^\circ$), AUC score at $0.5^\circ$, $1^\circ$, $2^\circ$ and avg.\ run-time (secs) on 1DSfM~\cite{wilson2014robust}.
    COLMAP is omitted since the reference comes from SfM.}
    \resizebox{0.75\columnwidth}{!}{
    \begin{tabular}{l c c c c c c} \toprule 
        & AVG ($^\circ$) & MED ($^\circ$) & AUC@$0.5^\circ$ &~~~~@$1^\circ~~~~$ &~~~~@$2^\circ~~~~$  & Time (s) \\ \midrule        
        LM 3DoF & 10.79 & 5.49 & \phantom{a}1.59 & \phantom{a}4.93 & 12.14 & 28.99 \\
        LM 1DoF & \phantom{a}3.94 & 1.76 & \phantom{a}4.01 & 11.77 & 26.68 & \phantom{a}\textbf{0.66} \\
        LAGO \cite{carlone2014fast} & \phantom{a}9.03 & 2.32 & \phantom{a}2.50 & \phantom{a}7.96 & 19.13 & \phantom{a}5.27 \\
        CPL-Sync \cite{fan2019efficient} & \phantom{a}4.82 & 1.88 & \phantom{a}3.45 & 10.24 & 23.41 & 62.17 \\
        Theia \cite{theia-manual} & \phantom{a}6.02 & 1.89 & \phantom{a}8.38 & 19.38 & 35.39 & 66.26 \\
        Theia$^\text{reg}$ \cite{crandall2012sfm} & \phantom{a}4.82 & 1.38 & \phantom{a}9.03 & 21.28 & 39.17 & 61.65 \\
        Proposed & \phantom{a}\textbf{1.54} & \textbf{0.82} & \textbf{11.02} & \textbf{29.41} & \textbf{54.76} & \phantom{a}4.38 \\
      \bottomrule
    \end{tabular}
    }
    \label{tbl:1dsfm}
\end{table}

%% file: tex/tbl/refine.tex
\begin{table}[t]
    \centering
    \caption{
    Average and median rot.\ errors ($^\circ$), AUC score at $0.5^\circ$, $1^\circ$ and $2^\circ$ on the 1DSfM dataset~\cite{wilson2014robust} with and without the gravity refinement proposed in Section~\ref{sec:refinement}.}
    \label{tbl:refinement}
    \resizebox{0.7\columnwidth}{!}{
    \begin{tabular}{l c c c c c} \toprule 
        & AVG ($^\circ$) & MED ($^\circ$) & AUC@$0.5^\circ$ &~~~~@$1^\circ~~~~$ &~~~~@$2^\circ~~~~$ \\ \midrule
        w/o refinement & 1.60 & 0.88 & \phantom{a}5.51 & 22.61 & 50.89 \\
        w/ refinement  & \textbf{1.54} & \textbf{0.82} & \textbf{11.02} & \textbf{29.41} & \textbf{54.76}\\
      \bottomrule
    \end{tabular}
    }
\end{table}

%% file: tex/text/conclusion.tex
\section{Conclusion}

We propose a principled algorithm that leverages the gravity direction to improve camera orientation estimation in rotation averaging for global SfM. 
The approach is based on circular statistics and simplifies to an iterative least-squares approach, providing similar convergence guarantees as ordinary linear regression.
By combining it with a robust loss function, we achieve significantly more accurate results than the baselines on four large-scale and real-world datasets.
It also supports scenarios where only a subset of cameras have known gravity.
Additionally, we propose a mechanism for refining gravity.
The proposed method runs several times faster than the widely used Theia library while reducing the error by a large margin.
Compared to LAGO, solving 1-DoF rotation averaging, the proposed algorithm is significantly more robust.
The code will be made public.

%% file: tex/text/appendix.tex
This supplementary material provides more results on the LaMAR datasets, 
an alternative solution to the problem based on changing the standard 3-DoF optimization, 
analysis of the period estimation, 
comparisons with deep learning-based alternatives,
results with gravity-aligned essential matrix,
implementation details for Theia baselines,
and detailed experiment results.



\section{Additional Results on LaMAR}
\input{tex/fig/result_lamar_hololens}

In the main paper, we consider gravity directions only in the data captured by smartphones to test the mixed scenario.
Here, we provide additional results for when HoloLens directions are used instead of the ones captured by smartphones. 
Moreover, we show results when both data sources have known gravity directions.

The results considering gravity only in the HoloLens sequences are in Fig.~\ref{fig:result_lamar_hololens}.
Planar pose graph optimization methods \cite{carlone2014fast, fan2019efficient} are excluded since they are not applicable in the mixed scenario.
A similar trend can be observed as in the case when only gravity captured by smartphones is known (Table 4 in the main paper).
The proposed method achieves higher AUC scores than Theia-based ones and is about 3-4 times faster.
While the proposed method has a higher average error than the algorithm implemented in Theia~\cite{theia-manual}, it is important to note that the average error is usually not representative in such problems since it is not a robust measure. 
In \textit{all} other metrics, the proposed method significantly outperforms Theia.
%
The proposed method is on par with COLMAP~\cite{schoenberger2016sfm}, having slightly lower AUC scores at $0.5^\circ, 1^\circ$ and a higher AUC score at $2^\circ$.
The proposed method also achieves lower mean and median errors and is faster by several orders of magnitude.
The cumulative distribution functions (CDF) of the absolute rotation errors are also shown in Fig. \ref{fig:result_lamar_hololens}.
The proposed method approaches the top left corner most, indicating its accuracy.

\input{tex/tbl/lamar_map_full}
The results for the case when gravity is available for both types of images are in Table~\ref{tbl:lamar_map_full}.
We exclude approximately $150$ images in total from this experiment for which the dataset does not provide gravity directions, and we keep only the largest connected component.
This is necessary to test \cite{carlone2014fast, fan2019efficient}.
Despite our tuning efforts, LAGO~\cite{carlone2014fast} and CPL-Sync~\cite{fan2019efficient} do not achieve reasonable accuracy on this dataset.
The proposed method achieves higher AUC scores at $0.5^\circ$, $1^\circ$, and $2^\circ$ compared with Theia-based methods and is about an order of magnitude faster.

The cumulative distribution functions (CDF) of the absolute rotation errors can be found in Figure~\ref{fig:result_lamar_sm}. 
The curve of the proposed method (red) is closest to the top left, implying that it obtains the most accurate estimation.


\input{tex/fig/result_full_sm}

\section{Alternative Representation}
\label{sec:alternative_representation}

As mentioned in the main text, an alternative way to achieve 1DoF optimization is to constrain the update step in the numerical optimization so that it only changes the rotation around the $y$-axis.
More precisely, in each optimization step, the updates on rotations are projected to be around the $y$-axis.
Though minimizing the same loss function as the proposed method, the alternative formulation requires solving a larger linear system. 
As a result, it achieves similar accuracy at the cost of a significantly longer runtime compared to the proposed method.
Furthermore, it is not applicable when gravity is partially known.
A detailed analysis of the relationship between the proposed method and such alternative formulation can be found in Section \ref{sec:spectrum_constraints}.
We will call this alternative formulation Proposed$^\text{alt}$. 

\input{tex/tbl/runtime}
The cumulative distribution functions (CDF) of the absolute rotation errors are shown in Fig.~\ref{fig:result_full_sm}.
As expected, the curves of the alternative formulation and the proposed methods nearly overlap, indicating that they obtain similar results.
The runtimes are summarized in Table~\ref{tbl:runtime}.
The proposed formulation, directly optimizing on the manifold, leads to a consistent 4-9 times speedup compared to the alternative method.



\section{Spectrum of Constraints}
\label{sec:spectrum_constraints}

In this section, we further inspect the formulation that incorporates gravity as an extra penalty term.
We show how it allows a spectrum of problems, varying from unconstrained to hard-constrained, and how it can be converted into the proposed method under a specific condition.

Recall the objective function with an extra penalty term (Eq. 21 in the main paper) is formulated as
\begin{equation}
\begin{split}
    \arg\min_{\{\bm{R}_i\}} \sum_{(i,j)\in \mathcal{E}} \rho\left(d(\bm{R}_j^\top \tilde{\bm{R}}_{ij} \bm{R}_i, \bm{I})^p\right)
    + \sum_{i} \lambda \rho\left(d(\theta \cdot \left(\bm{v}_i - (\bm{v}_i^\top\bm{g}_i)\bm{g}_i)\right)^p\right).
    \label{eq:R_err_penalty_full}
\end{split}
\end{equation}
Since $d_{\text{geod}}(\bm{R},\bm{S}) \approx \|\theta_R\bm{v}_R - \theta_S\bm{v}_S\|_2$, $\|\theta\cdot(\bm{v}_i - (\bm{v}_i^\top\bm{g}_i)\bm{g}_i)\|_2$ approximates the angle distance of $\bm{R}_i$ from the closest upright rotation.
Thus, the proposed formulation is similar to the penalty in \cite{crandall2012sfm}.

Parameter $\lambda$ in Eq.~\eqref{eq:R_err_penalty_full} controls the weighting between consistency with the relative pose and the measured gravity.
When $\lambda$ is 0, the second part of the objective function can be factored out, and Eq.~\eqref{eq:R_err_penalty_full} restores the original 3DoF optimization.
Naturally, as $\lambda$ increases, more emphasis is put on the consistency with measured gravity.
As a result, varying the level of $\lambda$ forms a spectrum of problems, ranging from soft-constrained to hard-constrained ones.
Worth noting is the case when $\lambda=\infty$.
The optimal solution in this case should possess the following property:
\begin{equation}
\left\|\theta \cdot\left(\bm{v}_i - (\bm{v}_i^\top\bm{g}_i)\bm{g}_i\right)\right\| = 0.
\end{equation}
To fulfill the above condition, the downward direction in the estimated rotation should be the same as the measured gravity.
If an iterative process solves the optimization, in each step, the updates are projected to act around the gravity direction.
As a result, the problem becomes the same as what is presented in Section~\ref{sec:alternative_representation}.

\input{tex/fig/sensitivity}

Besides varying the problem from non-constrained to hard-constrained ones, the above formulation also bridges the proposed method to the original 3DoF optimization.
When $\lambda = \infty$, suppose the rotation are pre-aligned that gravity direction $\bm{g}_i$ is $[0, 1, 0]^\top$.
Plugging this into the optimization process gives that $\bm{v}^t_x = \bm{v}^t_z = 0$.
This means the $x,z$ components remain 0 across the optimization process, thus the linear systems corresponding to these two rotation components become redundant.
Removing them results in a smaller linear system equivalent to the proposed system in the paper.
This observation also directly suggests the efficiency of our methods.
While our method reduces the number of equations in the linear system by $\sim 2m$ where $m$ is the number of pairs, formulation with soft penalty adds $\sim 2n$ additional equations where $n$ is the number of images.



\section{Sensitivity Analysis}
In this section, we analyze the sensitivity of the method in the presence of noise and outliers.
In particular, we report the AUC score as a function of gravity noise and the outlier ratio in the pre-estimated relative poses.
%

To analyze the sensitivity of the method to gravity noise, we alternate the level of noise in gravity by interpolating between measured and ground-truth gravity.
More concretely, it is formulated as 
\begin{equation}
    \bm{g}_i^{\alpha} = \frac{\alpha\cdot\bm{g}_i  + (1-\alpha)\cdot\bm{g}_i^{gt}}{\left\|\alpha\cdot\bm{g}_i  + (1-\alpha)\cdot\bm{g}_i^{gt}\right\|},
\end{equation}
where $\bm{g}_i$ is the measured gravity, $\bm{g}_i^{gt}$ is the ground truth gravity, and $\alpha$ controls the level of noise.
Such a setting allows for progressively mixing the noise in the measurements to the ground truth direction.
We vary $\alpha$ from 0.5 to 5, and results are summarized in Figure~\ref{fig:sensitivity_grav}.
The $x$-axis in the plot is the median gravity error, and $\alpha = 1$ corresponds to  $\sigma_\text{grav} = 0.54^\circ$, $0.20^\circ$ and $0.31^\circ$ respectively.
From the figure, it can be observed that methods incorporating gravity, \ie the proposed method and Theia with extra penalty term (Theia$^\text{reg}$), produce more accurate results compared to the original 3DoF optimization.
This finding further motivates the use of gravity in the rotation averaging problem.
Additionally, the proposed method obtains the most accurate estimations among all methods.
This holds until $\sim 0.7^\circ$ on the datasets we tested.
Such levels of accuracy can be satisfied by commercial sensors as the median gravity errors on EuRoC~\cite{burri2016euroc} and KITTI~\cite{geiger2013vision} are $0.54^\circ$ and $0.20^\circ$, respectively.
On LaMAR~\cite{sarlin2022lamar}, the median error of HoloLens images is $0.34^\circ$, and that of smartphone images is 0.26$^\circ$.

To analyze the sensitivity of the method to outliers, we conducted synthetic experiments with varying levels of outliers.
There, we apply the same grid setting as in the synthetic experiments in the main paper. 
For a brief recap, we assume that the cameras form a 2D grid, and each camera is connected with the closest 24 grid neighbors. 
In the experiments, the noise level for gravity prior is fixed to be $0.25^\circ$ which is comparable to that of gravity measurements.
As for the relative pose, we add zero-mean noise by applying random rotations with a standard deviation of $1^\circ$.
To inject outliers, a uniformly randomly drawn set of relative poses is replaced by random rotations.
Results are summarized in Figure~\ref{fig:sensitivity_outlier}.
The plot shows that the performance of the proposed method and baselines are similar when the relative poses are outlier-free.
As the outlier ratio increases, the accuracy in estimation for all methods degrades at different rates.
LAGO~\cite{carlone2014fast} is subject to drastic performance drop as the outlier ratios increase, validating the claim that its heuristic approach for fixing the period is not robust.
CPL-Sync~\cite{fan2019efficient} is similarly sensitive to outliers.
In contrast, Theia-based baselines and our method are robust in the presence of outliers.
Among all, the proposed method undergoes the least performance drop as the outlier ratio increases.
For the proposed method, the AUC score at $1^\circ$ only decreases by $11\%$ while those for Theia and Theia with penalty terms drop by $32\%$ and $24\%$, respectively.

\section{Effect of Iteration on Period Estimation}

LAGO~\cite{carlone2014fast} estimates periods for edges from a coarse initialization.
Such a heuristic approach often fails on different datasets.
For example, on the LaMAR dataset, the AUC score of LAGO at $1^\circ$ is only 0.94, while that of the proposed is 23.85, and on 1DSfM, the scores are 7.96 and 29.41, respectively.
We conduct experiments on 1DSfM~\cite{wilson2014robust}, EuRoC~\cite{burri2016euroc}, KITTI~\cite{geiger2013vision}, and LaMAR~\cite{sarlin2022lamar} to analyze the accuracy of period estimation (\textit{correct ratio}) for LAGO~\cite{carlone2014fast} and the proposed iterative solution.
Also, we analyze the relationship between period estimation and accuracy.

To calculate the ground truth periods, first, the ground truth rotations are globally aligned with the estimated results and are adjusted to be at most $\pi$ radian to the estimation.
Then, ground truth periods for edges are calculated using the same formulation as in LAGO~\cite{carlone2014fast} with the adjusted ground truth rotations.

\begin{figure}[t]
    \centering
    \subfloat[accuracy of period and camera pose estimation across iterations. $k$ is the period number] {
    \begin{tabular}{cc}
    \multicolumn{2}{c}{\includegraphics[width=0.65\columnwidth]{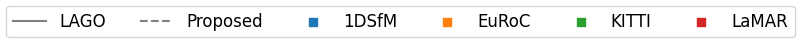}}\\
     \includegraphics[height=0.25\columnwidth]{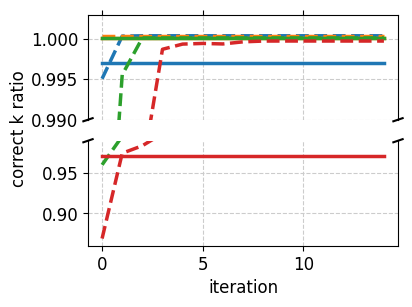} &
     \includegraphics[height=0.25\columnwidth]{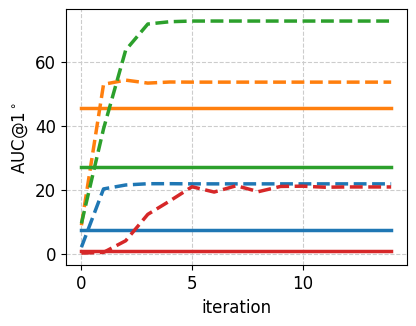} \\
    \end{tabular}
    \label{fig:ks_ite}
         }
     \subfloat[effect of fixing period]{
     \includegraphics[height=0.25\columnwidth]{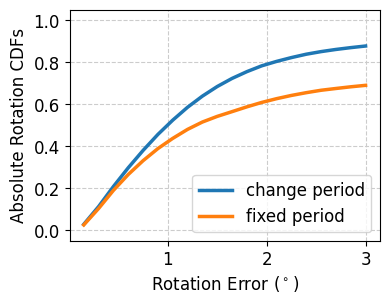}
     \label{fig:fix_period}
     }
    \caption{
    Period estimation can have a large impact on the accuracy of rotation averaging.
    }
\end{figure}

Results are in Fig.~\ref{fig:ks_ite}.
For iteration 0, the rotations for our method are initialized to be identity.
As LAGO is not an iterative method, we report its error as a constant number across iterations.
From the plot, one can observe that the \textit{correct ratio} is monotonically increasing for the proposed method.
Also, the convergence rate is generally fast and the proposed method can achieve almost 100\% \textit{correct ratio} within 5 iterations.
This indicates the effectiveness of the proposed mechanism.
While for LAGO~\cite{carlone2014fast}, though we can observe it obtain high \textit{correct ratio} on the EuRoC~\cite{burri2016euroc} and KITTI~\cite{geiger2013vision} datasets, which are within the domain the method was originally tailored for, its performance largely drops on other datasets.
As a result, LAGO fails.

The right plot of Fig. \ref{fig:ks_ite} shows the relationship between estimation accuracy and iteration.
By reading the \textit{correct ratio} from the left plot and the AUC score on the right plot, one can notice a strong positive correlation between these two variables.
Across iterations, the accuracy of the rotation orientation increases, leading to a more accurate estimation of periods.
These estimations can later contribute to a more accurate rotation estimation, and finally, the estimation converges to the desired result.


Fig.~\ref{fig:fix_period} shows the importance of alternating period estimation in the optimization as proposed in the main paper.
As from Fig.~\ref{fig:ks_ite}, the estimations of periods on EuRoC~\cite{burri2016euroc} and KITTI~\cite{geiger2013vision} are error-free, we only show the result on the LaMAR~\cite{sarlin2022lamar} dataset.
From the plot, it can be seen that the accuracy of the rotation averaging is largely affected by the accuracy of initial period estimation, even if a robust estimation scheme is deployed.
This indicates the necessity of deploying the circular regression.


\section{Comparison with Learning-based Methods}
\subsection{Unordered Image Collections}
\input{tex/tbl/ra_1dsfm}
In this section, we compare with the recent learning-based DMF-SYNCH~\cite{tejus2023rotation}, NeuRoRA~\cite{purkait2020neurora}, MSP~\cite{yang2021end} and PoGO-Net~\cite{li2021pogo}.
Results on the 1DSfM dataset are reported in Table \ref{tbl:ra_1dsfm}.
The results in the table are directly taken from the respective papers.
Given that we do not have a measured gravity direction, similarly as in the main paper, we take the gravity from the ground truth rotations and add zero-mean Gaussian noise with 0.5$^\circ$ standard deviation. 
The most accurate learning-based method is PoGO-Net. 
Over the scenes where results are provided (all, except for Gendarmenmarkt), the average rotation error of PoGO-Net is 0.96$^\circ$, while that of the proposed method is 0.79$^\circ$. 
On Gendarmenmarkt, PoGO-Net results are missing due to its high memory requirement, while the proposed method leads to significantly more accurate results than others.
Given that the gravity error is lower than 0.5$^\circ$ on most of the tested datasets, we expect even larger differences in practice. 


\subsection{Sequential Image Collections}

On the KITTI~\cite{geiger2013vision} and EuRoC~\cite{burri2016euroc} datasets, we failed to achieve reasonable results for DMF-SYNCH~\cite{tejus2023rotation}, NeuRoRA~\cite{purkait2020neurora} and MSP~\cite{yang2021end} even when using the code provided by the authors and retraining their models.
As DMF-SYNCH is a method based on matrix completion, it requires not too sparsely connected graphs, which sequential datasets generally do not satisfy.
For the shortest sequence 04 in KITTI, which only contains 277 images, DMF-SYNCH can achieve a median error at the level of 10$^\circ$. 
But for sequences with more than 1000 images, we failed to achieve a median error below 30$^\circ$.
Also, the parameters to learn grow quadratically with the number of images, making the method impractical for sequences with more than a few thousand images.

The sparsity also poses challenges to NeuRoRA~\cite{purkait2020neurora} and MSP~\cite{yang2021end}, which are graph-neural-network-based methods.
They contain two parts: a view-graph cleaning component and a rotation refinement component.
The first component in both works rejects outliers and refines relative rotations.
Then, NeuRoRA and MSP initialize global rotation estimation via a minimal spanning tree on the input view graph.
After coarse initialization, they both proceed with a network called FineNet, which refines the rotations.

We attempted to run NeuRoRA~\cite{purkait2020neurora} in different ways. 
We tried the weights provided by the authors, finetuning the network on 1DSfM, finetuning on KITTI, and finetuning on EuRoC.
All such tests led to inaccurate results.
To eliminate the effect of the potentially failing first step (\ie, view-graph cleaning), we tried providing the ground truth outlier information as initialization for the refinement module.
Since the remaining relative poses are generally accurate, the obtained initialization is reasonable, with median errors generally below 3$^\circ$ for EuRoC and KITTI sequences.
Despite this accurate initialization, FineNet consistently (with all training strategies) reduced the accuracy instead of improving it.
%
We conclude that these two methods are not suitable for the rotation averaging on KITTI, EuRoC, and LaMAR datasets featuring long sequential image streams.

\input{tex/tbl/KITTI_pogonet}
As PoGO-Net~\cite{li2021pogo} has no official code available, we report the errors on the KITTI dataset in Table~\ref{tbl:KITTI_pogonet} as provided in the original paper. 
The proposed method substantially improves upon PoGO-Net on both sequences. 

\section{Results with Gravity-Aligned Essential Matrix}
The results when using 3-DoF and 5-DoF relative poses as input are reported in Table~\ref{tbl:3dof_E}.
The 3-DoF essential matrix solver is incorporated within the same LO-RANSAC framework used in other experiments to estimate gravity-aligned essential matrix.
We allow for full 5 DoF optimization in the essential matrix estimation.
We can see that the gravity-aligned essential matrix improves almost all results, except for Theia’s AVG error on KITTI.
The proposed method is still significantly more accurate than Theia~\cite{theia-manual}.
\input{tex/tbl/3dof_E}

\section{Implementation Details for Theia Baselines}
For the ease of implementation of Theia with regularization terms, we pre-align the camera rotations with the gravity prior.
Thus, the estimated rotations are around $(0, 1, 0)^\top$.
In this case, the first and the last term for camera rotations should be 0.
We can directly penalize these two terms as the deviation from the prior.
After appending these linear systems to the system, the problem can be solved as in Theia~\cite{theia-manual}.

\section{Detailed Experiment Results}
Per sequence result for EuRoC~\cite{burri2016euroc} and KITTI~\cite{geiger2013vision} for the proposed method are summarized in Table~\ref{tbl:full_euroc}, \ref{tbl:full_kitti}.
From the tables, one can observed that the proposed method achieves accurate rotation estimation consistently across the datasets.

\input{tex/tbl/sm_full}


%% file: tex/fig/result_lamar_hololens.tex
\begin{figure}[h]
    \centering
    \includegraphics[width=0.4\columnwidth]{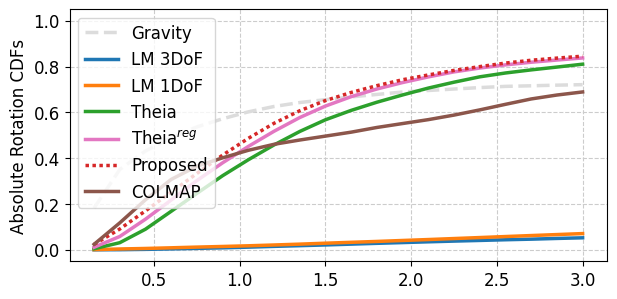}
    \resizebox{0.6\columnwidth}{!}{
    \begin{tabular}{l c c c c c c} \toprule 
        & AVG ($^\circ$) & MED ($^\circ$) & AUC@$0.5^\circ$ & @$1^\circ$ & @$2^\circ$  & Time (h) \\ \midrule
        LM 3DoF & 52.30 & 32.36 & 0.07 & \phantom{a}0.35 & \phantom{a}1.26 & 4.08 $\times 10^{-2}$ \\
        LM 1DoF & 44.90 & 31.61 & 0.28 & \phantom{a}0.73 & \phantom{a}1.81 & 1.69 $\times 10^{-2}$ \\
        Theia \cite{theia-manual} & \phantom{a}\textbf{3.46} & \phantom{a}1.28 & 3.28 & 13.92 & 34.62 & 3.88 $\times 10^{-2}$ \\
        Theia$^\text{reg}$ \cite{crandall2012sfm} & \phantom{a}8.14 & \phantom{a}1.13 & 5.36 & 17.65 & 39.42 & 4.31 $\times 10^{-2}$ \\
        Proposed & \phantom{a}7.19 & \phantom{a}\textbf{1.06} & 7.63 & 20.46 & \textbf{42.02} & \textbf{1.21} $\times 10^{-2}$ \\
        \hdashline
        COLMAP \cite{schoenberger2016sfm} & 34.18 & \phantom{a}1.42 & \textbf{9.63} & \textbf{22.59} & 36.09 & 98.52 \\
      \bottomrule
    \end{tabular}
    }
    \caption{Results for on LaMAR~\cite{sarlin2022lamar}. Only images taken by Hololens data are with gravity, thus  \cite{carlone2014fast, fan2019efficient} are not applicable.}
    
    \label{fig:result_lamar_hololens}
\end{figure}

%% file: tex/tbl/lamar_map_full.tex
\begin{table}[h]
    \centering
    \caption{Average and median rotation errors ($^\circ$), the AUC score at $0.5^\circ$, $1^\circ$, and $2^\circ$ and average run-time (hour) on LaMAR~\cite{sarlin2022lamar}. Images without gravity are excluded.
     }
    \resizebox{0.6\columnwidth}{!}{
    \begin{tabular}{l c c c c c c} \toprule 
        & AVG ($^\circ$) & MED ($^\circ$) & AUC@$0.5^\circ$ & @$1^\circ$ & @$2^\circ$  & Time (h) \\ \midrule
        LM 3DoF & 52.31 & 32.20 & \phantom{a}0.08 & \phantom{a}0.36 & \phantom{a}1.25 & 4.05 $\times 10^{-2}$ \\
        LM 1DoF & 46.25 & 35.76 & \phantom{a}0.26 & \phantom{a}0.78 & \phantom{a}1.98 & \textbf{1.28 $\times 10^{-3}$} \\
        LAGO \cite{carlone2014fast} & 51.28 & 37.17 & \phantom{a}0.33 & \phantom{a}0.94 & \phantom{a}2.22 & 5.18 $\times 10^{-3}$ \\
        CPL-Sync \cite{fan2019efficient} & 11.57 & \phantom{a}7.95 & \phantom{a}0.92 & \phantom{a}2.63 & \phantom{a}6.27 & 3.09 $\times 10^{-3}$ \\
        Theia \cite{theia-manual} & \phantom{a}3.52 & \phantom{a}1.28 & \phantom{a}3.26 & 13.82 & 34.36 & 4.03 $\times 10^{-2}$ \\
        Theia$^\text{reg}$ \cite{crandall2012sfm} & \phantom{a}4.26 & \phantom{a}1.04 & \phantom{a}7.12 & 20.91 & 41.99 & 3.79 $\times 10^{-2}$ \\
        Proposed & \phantom{a}\textbf{2.90} & \phantom{a}0.94 & \phantom{a}9.32 & 23.85 & \textbf{46.24} & 3.88 $\times 10^{-3}$ \\
        \hdashline
        COLMAP \cite{schoenberger2016sfm} & \phantom{a}9.68 & \phantom{a}\textbf{0.88} & \textbf{12.71} & \textbf{28.38} & 44.22 & 98.52 \\
      \bottomrule
    \end{tabular}
    }
    \label{tbl:lamar_map_full}
\end{table}

%% file: tex/fig/result_full_sm.tex
\begin{figure*}
    \centering
     \includegraphics[width=0.9\textwidth]{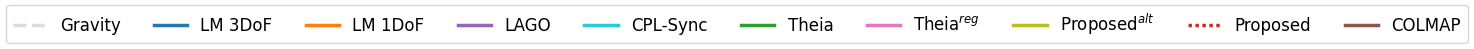}\\
     \begin{subfigure}[b]{0.255\textwidth}
         \centering
         \includegraphics[width=\textwidth]{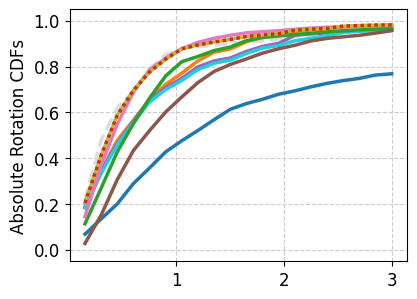}
         \caption{EuRoC \cite{burri2016euroc}}
         \label{fig:result_euroc_sm}
     \end{subfigure}
     \begin{subfigure}[b]{0.23\textwidth}
         \centering
         \includegraphics[width=\textwidth]{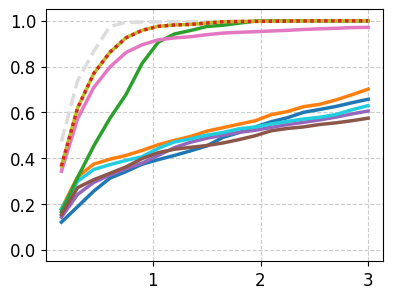}
         \caption{KITTI \cite{geiger2013vision}}
         \label{fig:result_kitti_sm}
     \end{subfigure}
     \begin{subfigure}[b]{0.23\textwidth}
         \centering
         \includegraphics[width=\textwidth]{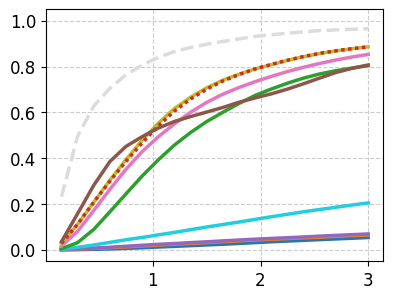}
         \caption{LaMAR \cite{sarlin2022lamar}}
         \label{fig:result_lamar_sm}
     \end{subfigure}
     \begin{subfigure}[b]{0.23\textwidth}
         \centering
         \includegraphics[width=\textwidth]{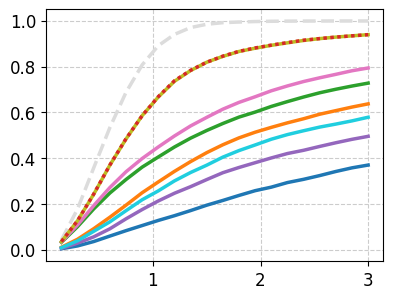}
         \caption{1DSfM \cite{wilson2014robust}}
         \label{fig:result_1dsfm_sm}
     \end{subfigure}
     \vspace{-5px}
    \caption{ 
    The cumulative distribution functions (CDFs) of the absolute rotation errors ($^\circ$). Estimated by the Levenberg-Marquardt~\cite{more2006levenberg} method solving the 3-DoF (LM 3DoF) and 1-DoF (LM 1DoF) problems, by LAGO~\cite{carlone2014fast}, by CPL-Sync~\cite{fan2019efficient}, by the rotation averaging in the Theia library~\cite{theia-manual,chatterjee2013efficient}, by Theia with an additional penalty term (Theia$^\text{reg}$ \cite{crandall2012sfm}), by projecting updates to y-axis (Proposed$^\text{alt}$), by COLMAP~\cite{schoenberger2016sfm}, and by the proposed method.
    %
    Curve "Gravity" stands for the approximate upper bound achievable by using gravity direction.
    %
    A method being accurate is interpreted by its curve close to the top-left corner.
    }
    \label{fig:result_full_sm}
\end{figure*}

%% file: tex/tbl/runtime.tex
\begin{table}[h]
    \centering
    \caption{
    Runtime (s) for the projected method and direct optimization on the manifold. The latter is 4-9 times faster.}
    \resizebox{0.6\columnwidth}{!}{
    \begin{tabular}{lcccc}
    \toprule
    & EuRoC~\cite{burri2016euroc} & KITTI~\cite{geiger2013vision} & LaMAR~\cite{sarlin2022lamar} & 1DSfM~\cite{wilson2014robust} \\
    \midrule
    Proposed$^\text{alt}$ & 0.84 & 0.74 & 113.76 & 26.54 \\
    Proposed & \textbf{0.20} & \textbf{0.20} & \phantom{a}\textbf{13.98} & \phantom{a}\textbf{4.38} \\
    \bottomrule
    \end{tabular}
    }
    \label{tbl:runtime}
\end{table}

%% file: tex/fig/sensitivity.tex
\begin{figure*}[t]
    \centering
    \subfloat[median gravity error vs AUC@1$^\circ$. $\alpha=1$ corresponds to $\sigma_{\text{grav}}$ equals 0.54$^\circ$, 0.20$^\circ$, and 0.31$^\circ$ respectively]{%
         \includegraphics[width=0.23\textwidth]{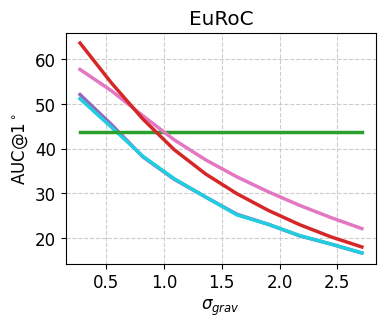}
         \includegraphics[width=0.23\textwidth]{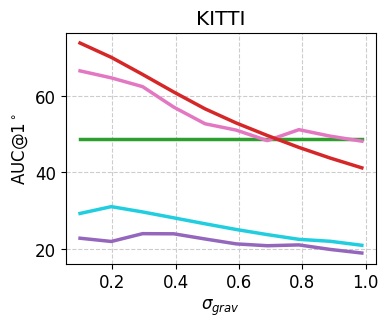}
         \includegraphics[width=0.23\textwidth]{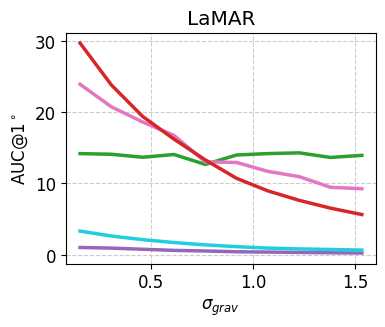}
         \label{fig:sensitivity_grav}
     }
    \subfloat[outlier vs AUC@1$^\circ$] {
         \includegraphics[width=0.23\textwidth]{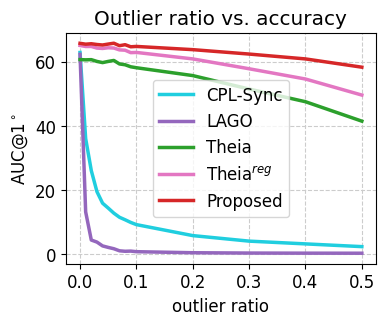}
         \label{fig:sensitivity_outlier}
     }
    \vspace{-5px} 
    \caption{Sensitivity analysis w.r.t.\ gravity noise and outlier ratio in the measured relative poses.
    For the left plots, gravity are synthesized by interpolating between noisy measurements and ground truth with different weights.
    For the right plot, synthetic experiments with grid setting are conducted. In these experiments, different level of outlier are tested with fixed noise level on gravity and relative poses.
    }
    \label{fig:sensitivity}
\end{figure*}

%% file: tex/tbl/ra_1dsfm.tex
\begin{table}[t]
    \centering
\caption{Results on 1DSfM dataset \cite{wilson2014robust}. Columns marked with $^\dagger$ are from the respective paper.
We use ground truth with 0.5$^\circ$ standard deviation error as gravity.
}
    \resizebox{0.6\columnwidth}{!}{
    \begin{tabular}{l|cccc|c}
    \hline
     & DMF-SYNCH$^\dagger$ & NeuRoRa$^\dagger$ & MSP$^\dagger$ & PoGO-Net$^\dagger$ & \multirow{2}{*}{Proposed} \\
     & \cite{tejus2023rotation} & \cite{purkait2020neurora} & \cite{yang2021end} & \cite{li2021pogo} \\
    \hline
    ALM & \phantom{a}1.2 & 1.2 & 1.07 & \textbf{0.85} & 0.90 \\
    ELS & \phantom{a}0.8 & 0.6 & 0.83 & \textbf{0.43} & 0.52 \\
    GDM & 10.5 & 2.9 & 3.69 & - & \textbf{0.96} \\
    MDR & \phantom{a}2.3 & 1.1 & 1.09 & 0.96 & \textbf{0.67} \\
    MND & \phantom{a}0.6 & 0.6 & 0.50 & \textbf{0.37} & 0.51 \\
    NYC & \phantom{a}1.8 & 1.1 & 1.12 & \textbf{0.88} & 1.03 \\
    PDP & \phantom{a}1.0 & 0.7 & 0.76 & 0.81 & \textbf{0.66} \\
    PIC & - & 1.9 & 1.80 & 1.75 & \textbf{0.80} \\
    ROF & \phantom{a}1.8 & 1.3 & 1.19 & \textbf{0.69} & 0.71 \\
    TOL & \phantom{a}2.7 & 1.4 & 1.25 & \textbf{0.43} & 0.64 \\
    TFG & - & 2.2 & - & 1.70 & \textbf{0.86} \\
    USQ & \phantom{a}4.4 & 2.0 & 1.85 & \textbf{1.25} & 1.52 \\
    VNC & \phantom{a}1.6 & 1.5 & 1.10 & 1.44 & \textbf{0.74} \\
    YKM & \phantom{a}1.7 & 0.9 & 0.91 & \textbf{0.72} & \textbf{0.72} \\
    
    \hline
    \end{tabular}
}
\label{tbl:ra_1dsfm}
\end{table}

%% file: tex/tbl/KITTI_pogonet.tex
\begin{table}[]
    \centering  
    \caption{Mean rotation error ($^\circ$) on sequence 02 and 08 in KITTI~\cite{geiger2013vision}. Results with $^\dagger$ are taken from the original paper.
    }
    \resizebox{0.37\columnwidth}{!}{
    \begin{tabular}{l c c} \toprule 
        & PoGO-Net~\cite{li2021pogo}$^\dagger$ & Proposed \\ \midrule

        KITTI-02 & 1.08 & \textbf{0.46} \\
        KITTI-08 & 2.17 & \textbf{0.53} \\
      \bottomrule
    \end{tabular}
    }
    \label{tbl:KITTI_pogonet}
\end{table}

%% file: tex/tbl/3dof_E.tex
\begin{table}[h]
    \centering
    \caption{Results with relative poses from both 5-DoF solvers and 3-DoF solvers for EuRoC and KITTI dataset.}
    \resizebox{0.7\columnwidth}{!}{
    \begin{tabular}{c l c c c c c c} \toprule 
        ~~~~~~~~&&& AVG ($^\circ$) & MED ($^\circ$) & AUC@$0.5^\circ$ & ~~~~@$1^\circ$~~~~& ~~~~@$2^\circ$~~~~ \\ \midrule
\multirow{4}{*}{\rotatebox{90}{\textit{EuRoC}}} &
\multirow{2}{*}{Proposed} & 5DoF & 0.65 & 0.35 & 32.68 & 53.89 & 72.11 \\
&& 3DoF & \textbf{0.62} & \textbf{0.34} & \textbf{33.94} & \textbf{56.64} & \textbf{74.12} \\
\cdashline{2-8} 
& \multirow{2}{*}{Theia} & 5DoF & 0.88 & 0.54 & 22.01 & 43.30 & 64.92 \\
&& 3DoF & \textbf{0.77} & \textbf{0.48} & \textbf{23.56} & \textbf{47.05} & \textbf{68.80} \\
\midrule
\multirow{4}{*}{\rotatebox{90}{\textit{KITTI}}} &
\multirow{2}{*}{Proposed} & 5DoF & \textbf{0.39} & 0.29 & 48.71 & 69.79 & 83.86  \\ 
&& 3DoF & \textbf{0.39} & \textbf{0.27} & \textbf{50.69} & \textbf{70.51} & \textbf{83.98} \\
\cdashline{2-8} 
&\multirow{2}{*}{Theia} & 5DoF & \textbf{0.68} & 0.64 & 25.96 & 47.48 & 71.77 \\
&& 3DoF & 2.52 & \textbf{0.54} & \textbf{28.79} & \textbf{52.58} & \textbf{72.92} \\
      \bottomrule
    \end{tabular}
    }
    \label{tbl:3dof_E}
\end{table}

%% file: tex/tbl/sm_full.tex
\begin{table}[h]
    \centering
    \caption{
    Full results of the proposed method for EuRoC~\cite{burri2016euroc}
    }
    \resizebox{\columnwidth}{!}{
    \begin{tabular}{lccccccccccc}
\toprule
& ~MH\_01~ & ~MH\_02~ & ~MH\_03~ & ~MH\_04~ & ~MH\_05~ & ~V1\_01~ & ~V1\_02~ & ~V1\_03~ & ~V2\_01~ & ~V2\_02~ & ~V2\_03~ \\
\midrule
mean & 0.78 & 0.21 & 0.20 & 0.46 & 0.18 & 1.79 & 0.32 & 0.58 & 0.48 & 0.69 & 0.63 \\
median & 0.26 & 0.21 & 0.15 & 0.35 & 0.16 & 1.48 & 0.32 & 0.56 & 0.48 & 0.58 & 0.57 \\
    \bottomrule
    \end{tabular}}
    \label{tbl:full_euroc}
\end{table}
\begin{table}[h]
    \centering
    \caption{
    Full results of the proposed method for KITTI~\cite{geiger2013vision}
    }
    \resizebox{0.7\columnwidth}{!}{
    \begin{tabular}{lccccccccc}
\toprule
& ~~~01~~~ & ~~~02~~~ & ~~~04~~~ & ~~~05~~~ & ~~~06~~~ & ~~~07~~~ & ~~~08~~~ & ~~~09~~~ & ~~~10~~~ \\
\midrule
mean & 0.35 & 0.46 & 0.07 & 0.39 & 0.20 & 0.38 & 0.53 & 0.23 & 0.22 \\
median & 0.32 & 0.40 & 0.06 & 0.35 & 0.15 & 0.24 & 0.30 & 0.19 & 0.20 \\
    \bottomrule
    \end{tabular}}
    \label{tbl:full_kitti}
\end{table}